\documentclass[twoside,11pt]{article}

\usepackage{jmlr2e}
\usepackage{tabularx,booktabs,multirow}

\ShortHeadings{Meta-Learning: A Survey}{Joaquin Vanschoren}

\begin{document}

\title{Meta-Learning: A Survey}
\author{\name Joaquin Vanschoren \email j.vanschoren@tue.nl\\
       \addr Eindhoven University of Technology\\
       5600MB Eindhoven, The Netherlands}


\maketitle

\begin{abstract}
Meta-learning, or \emph{learning to learn}, is the science of systematically observing how different machine learning approaches perform on a wide range of learning tasks, and then learning from this experience, or \emph{meta-data}, to learn new tasks much faster than otherwise possible. Not only does this dramatically speed up and improve the design of machine learning pipelines or neural architectures, it also allows us to replace hand-engineered algorithms with novel approaches learned in a data-driven way. In this chapter, we provide an overview of the state of the art in this fascinating and continuously evolving field.
\end{abstract}

\section{Introduction}
\label{sec:introduction}
When we learn new skills, we rarely - if ever - start from scratch. We start from skills learned earlier in related tasks, reuse approaches that worked well before, and focus on what is likely worth trying based on experience \citep{lake2017building}. 
With every skill learned, learning new skills becomes easier, requiring fewer examples and less trial-and-error. In short, we \emph{learn how to learn} across tasks. Likewise, when building machine learning models for a specific task, we often build on experience with related tasks, or use our (often implicit) understanding of the behavior of machine learning techniques to help make the right choices.  

The challenge in meta-learning is to learn from prior experience in a systematic, data-driven way.  First, we need to collect \emph{meta-data} that describe prior learning tasks and previously learned models. They comprise the exact \emph{algorithm configurations} used to train the models, including hyperparameter settings, pipeline compositions and/or network architectures, the resulting \emph{model evaluations}, such as accuracy and training time, the learned model parameters, such as the trained weights of a neural net, as well as measurable properties of the task itself, also known as \emph{meta-features}. Second, we need to \emph{learn} from this prior meta-data, to extract and transfer knowledge that guides the search for optimal models for new tasks. This chapter presents a concise overview of different meta-learning approaches to do this effectively. 

The term \emph{meta-learning} covers any type of learning based on prior experience with other tasks. The more \emph{similar} those previous tasks are, the more types of meta-data we can leverage, and defining task similarity will be a key overarching challenge. Perhaps needless to say, there is no free lunch \citep{wolpert+96,giraud2005toward}. When a new task represents completely unrelated phenomena, or random noise, leveraging prior experience will not be effective. Luckily, in real-world tasks, there are plenty of opportunities to learn from prior experience.

In the remainder of this chapter, we categorize meta-learning techniques based on the type of meta-data they leverage, from the most general to the most task-specific. First, in Section \ref{sec:evaluations}, we discuss how to \emph{learn purely from model evaluations}. These techniques can be used to recommend generally useful configurations and configuration search spaces, as well as transfer knowledge from \emph{empirically similar} tasks. In Section \ref{sec:metafeatures}, we discuss how we can \emph{characterize} tasks to more explicitly express task similarity and build meta-models that learn the relationships between data characteristics and learning performance. Finally, Section \ref{sec:modeltransfer} covers how we can \emph{transfer trained model parameters} between tasks that are inherently similar, e.g. sharing the same input features, which enables transfer learning \citep{pan2010survey} and few-shot learning \citep{ravi2016optimization}.

Note that while \emph{multi-task learning} \citep{Caruana97} (learning multiple related tasks simultaneously) and \emph{ensemble learning} \citep{Dietterich:2000p4081} (building multiple models on the same task), can often be meaningfully combined with meta-learning systems, they do not in themselves involve learning from prior experience on other tasks. 


\section{Learning from Model Evaluations}
\label{sec:evaluations}
Consider that we have access to prior tasks $t_{j} \in T$, the set of all known tasks, as well as a set of learning algorithms, fully defined by their \emph{configurations} $\theta_{i} \in \Theta$; here $\Theta$ represents a discrete, continuous, or mixed configuration space which can cover hyperparameter settings, pipeline components and/or network architecture components. $\mathbf{P}$ is the set of all prior scalar evaluations $P_{i,j} = P(\theta_{i},t_{j})$ of configuration $\theta_{i}$ on task $t_{j}$, according to a predefined evaluation measure, e.g. accuracy, and model evaluation technique, e.g. cross-validation. $\mathbf{P}_{new}$ is the set of known evaluations $P_{i,new}$ on a new task $t_{new}$. We now want to train a \emph{meta-learner} $L$ that 
predicts recommended configurations $\Theta^{*}_{new}$ for a new task $t_{new}$. The meta-learner is trained on meta-data $\mathbf{P} \cup \mathbf{P}_{new}$. $\mathbf{P}$ is usually gathered beforehand, or extracted from meta-data repositories \citep{Vanschoren2014, vanschoren2012experiment}. $\mathbf{P}_{new}$ is learned by the meta-learning technique itself in an iterative fashion, sometimes \emph{warm-started} with an initial $\mathbf{P}_{new}^{'}$ generated by another method.


\subsection{Task-Independent Recommendations}
\label{sec:rankings}
First, imagine not having access to any evaluations on $t_{new}$, hence $\mathbf{P}_{new} = \varnothing$. We can then still learn a function $f: \Theta \times T \rightarrow \{\theta^{*}_{k}\}$, $k = 1 .. K$, yielding a set of recommended configurations \emph{independent} of $t_{new}$. These $\theta^{*}_{k}$ can then be evaluated on $t_{new}$ to select the best one, or to warm-start further optimization approaches, such as those discussed in Section \ref{sec:hptransfer}. 

Such approaches often produce a ranking, i.e. an \emph{ordered} set $\theta^{*}_{k}$. This is typically done by discretizing $\Theta$ into a set of candidate configurations $\theta_{i}$, also called a \emph{portfolio}, evaluated on a large number of tasks $t_{j}$. We can then build a ranking per task, for instance using \emph{success rates}, \emph{AUC}, or \emph{significant wins}~\citep{brazdil+03,demsar06,leite+12}. However, it is often desirable that equally good but faster algorithms are ranked higher, and multiple methods have been proposed to trade off accuracy and training time \citep{brazdil+03,vanrijn2015}. Next, we can aggregate these single-task rankings into a \emph{global ranking}, for instance by computing the average rank \citep{Lin2010,Abdulrahman+18} across all tasks. 
When there is insufficient data to build a global ranking, one can recommend \emph{subsets of configurations} based on the best known configurations for each prior task \citep{Todorovski:1999p5595,kalousis02}, or return \emph{quasi-linear rankings} \citep{cook+96}. 

To find the best $\theta^{*}$ for a task $t_{new}$, never before seen, a simple anytime method is to select the top-$K$ configurations \citep{brazdil+03}, going down the list and evaluating each configuration on $t_{new}$ in turn. This evaluation can be halted after a predefined value for $K$, a time budget, or when a sufficiently accurate model is found. In time-constrained settings, it has been shown that multi-objective rankings (including training time) converge to near-optimal models much faster \citep{Abdulrahman+18,vanrijn2015}, and provide a strong baseline for algorithm comparisons \citep{Abdulrahman+18,leite+12}. 

A very different approach to the one above is to first fit a differentiable function $f_{j}(\theta_{i}) = P_{i,j}$ on all prior evaluations of a specific task $t_{j}$, and then use gradient descent to find an optimized configuration $\theta^{*}_{j}$ per prior task \citep{Wistuba2015}. Assuming that some of the tasks $t_{j}$ will be similar to $t_{new}$, those $\theta^{*}_{j}$ will be useful for warm-starting Bayesian optimization approaches. 

\subsection{Configuration Space Design}
Prior evaluations can also be used to learn a better \emph{configuration space} $\Theta^{*}$. While again independent from $t_{new}$, 
this can radically speed up the search for optimal models, since only the more relevant regions of the configuration space are explored. This is critical when computational resources are limited, and proves to be an important factor in practical comparisons of AutoML systems \citep{RECIPE}.

First, in the functional ANOVA \citep{fANOVA} approach, hyperparameters are deemed important if they explain most of the variance in algorithm performance on a given task. \citet{vanRijn2018b} evaluated this technique using 250,000 OpenML experiments with 3 algorithms across 100 datasets. 

An alternative approach is to first \emph{learn} an optimal hyperparameter default setting, and then define hyperparameter importance as the \emph{performance gain} that can be achieved by tuning the hyperparameter instead of leaving it at that default value. Indeed, even though a hyperparameter may cause a lot of variance, it may also have one specific setting that always results in good performance. \citet{Probst2018} do this using about 500,000 OpenML experiments on 6 algorithms and 38 datasets. Default values are learned \emph{jointly} for all hyperparameters of an algorithm by first training surrogate models for that algorithm for a large number of tasks. Next, many configurations are sampled, and the configuration that minimizes the average risk across all tasks is the recommended default configuration.  Finally, the importance (or \emph{tunability}) of each hyperparameter is estimated by observing how much improvement can still be gained by tuning it. 

\citet{Weerts2018} learn defaults \emph{independently} from other hyperparameters, and defined as the configurations that occur most frequently in the top-$K$ configurations for every task. In the case that the optimal default value depends on meta-features (e.g. the number of training instances or features), simple functions are learned that include these meta-features. Next, a statistical test defines whether a hyperparameter can be safely left at this default, based on the \emph{performance loss} observed when \emph{not} tuning a hyperparameter (or a set of hyperparameters), while all other parameters are tuned. This was evaluated using 118,000 OpenML experiments with 2 algorithms (SVMs and Random Forests) across 59 datasets.


\subsection{Configuration Transfer}
\label{sec:hptransfer}
If we want to provide recommendations for a specific task $t_{new}$, we need additional information on how similar $t_{new}$ is to prior tasks $t_j$. One way to do this is to evaluate a number of recommended (or potentially random) configurations on $t_{new}$, yielding new evidence $\mathbf{P}_{new}$. If we then observe that the evaluations $P_{i,new}$ are similar to $P_{i,j}$, then $t_{j}$ and $t_{new}$ can be considered intrinsically similar, based on empirical evidence. We can include this knowledge to train a meta-learner that predicts a recommended set of configurations $\Theta^{*}_{new}$ for $t_{new}$.
Moreover, every selected $\theta^{*}_{new}$ can be evaluated and included in $\mathbf{P}_{new}$, repeating the cycle and collecting more empirical evidence to learn which tasks are similar to each other. 

\subsubsection{Relative Landmarks}
A first measure for task similarity considers the relative (pairwise) performance differences, also called \emph{relative landmarks}, $RL_{a,b,j} = P_{a,j} - P_{b,j}$ between two configurations $\theta_{a}$ and $\theta_{b}$ on a particular task $t_{j}$ \citep{Furnkranz:2001p1278}. \emph{Active testing}~\citep{leite+12} leverages these as follows: it warm-starts with the globally best configuration (see Section \ref{sec:rankings}), calls it $\theta_{best}$, and proceeds in a tournament-style fashion. In each round, it selects the `competitor' $\theta_{c}$ that most convincingly outperforms $\theta_{best}$ on similar tasks. It deems tasks to be similar if the relative landmarks of all evaluated configurations are similar, i.e., if the configurations perform similarly on both $t_{j}$ and $t_{new}$ then the tasks are deemed similar. Next, it evaluates the competitor $\theta_{c}$, yielding $P_{c,new}$, updates the task similarities, and repeats. A limitation of this method is that it can only consider configurations $\theta_{i}$ that were evaluated on many prior tasks.

\subsubsection{Surrogate Models}
A more flexible way to transfer information is to build \emph{surrogate models} $s_{j}(\theta_{i}) = P_{i,j}$ for all prior tasks $t_{j}$, trained using all available $\mathbf{P}$. One can then define task similarity in terms of the error between $s_{j}(\theta_{i})$ and $P_{i,new}$: if the surrogate model for $t_{j}$ can generate accurate predictions for $t_{new}$, then those tasks are intrinsically similar. This is usually done in combination with Bayesian optimization \cite{rasmussen2004gaussian} to determine the next $\theta_{i}$.

\citet{Wistuba2018} train surrogate models based on Gaussian Processes (GPs) for every prior task, plus one for $t_{new}$, and combine them into a weighted, normalized sum, with the (new) mean $\mu$ defined as the weighted sum of the individual $\mu_{j}$'s (obtained from prior tasks $t_{j}$). The weights of the $\mu_{j}$'s are computed using the Nadaraya-Watson kernel-weighted average, where each task is represented as a vector of relative landmarks, and the Epanechnikov quadratic kernel \citep{nadaraya1964estimating} is used to measure the similarity between the relative landmark vectors of $t_{j}$ and $t_{new}$. The more similar $t_{j}$ is to $t_{new}$, the larger the weight $s_{j}$, increasing the influence of the surrogate model for $t_{j}$.

\citet{Feurer2018} propose to combine the predictive distributions of the individual Gaussian processes, which makes the combined model a Gaussian process again. The weights are computed following the agnostic Bayesian ensemble of \citet{lacoste2014agnostic}, which weights predictors according to an estimate of their generalization performance.

Meta-data can also be transferred in the acquisition function rather than the surrogate model \citep{Wistuba2018}. The surrogate model is only trained on $P_{i,new}$, but the next $\theta_{i}$ to evaluate is provided by an acquisition function which is the weighted average of the expected improvement \citep{jones1998efficient} on $P_{i,new}$ and the predicted improvements on all prior $P_{i,j}$. The weights of the prior tasks can again be defined via the accuracy of the surrogate model or via relative landmarks. The weight of the expected improvement component is gradually increased with every iteration as more evidence $P_{i,new}$ is collected.

\subsubsection{Warm-Started Multi-task Learning}
Another approach to relate prior tasks $t_{j}$ is to learn a joint task representation using $\mathbf{P}$. \citet{perrone2017multiple} train task-specific Bayesian linear regression \citep{bishop2006pattern} surrogate models $s_{j}(\theta_{i})$ and combine them in a feedforward Neural Network $NN(\theta_{i})$ which learns a joint task representation that can accurately predict $P_{i,new}$. The surrogate models are pre-trained on OpenML meta-data to provide a warm-start for optimizing $NN(\theta_{i})$ in a multi-task learning setting. Earlier work on multi-task learning \citep{swersky2013multi} assumed that we already have a set of `similar' source tasks $t_{j}$. It transfers information between these $t_{j}$ and $t_{new}$ by building a joint GP model for Bayesian optimization that learns and exploits the exact relationship between the tasks. Learning a joint GP tends to be less scalable than building one GP per task, though. \citet{Springenberg2016} also assume that the tasks are related and similar, but learns the relationship between tasks during the optimization process using Bayesian Neural Networks. As such, their method is somewhat of a hybrid of the previous two approaches. \citet{Golovin2017} assume a sequence order (e.g., time) across tasks. It builds a stack of GP regressors, one per task, training each GP on the residuals relative to the regressor below it. Hence, each task uses the tasks before it as its priors. 
 
\subsubsection{Other Techniques}
Multi-armed bandits \citep{robbins1985some} provide yet another approach to find the source tasks $t_{j}$ most related to $t_{new}$ \citep{Ramachandran2018}. In this analogy, each $t_{j}$ is one arm, and the (stochastic) reward for selecting (pulling) a particular prior task (arm) is defined in terms of the error in the predictions of a GP-based Bayesian optimizer that models the prior evaluations of $t_{j}$ as noisy measurements and combines them with the existing evaluations on $t_{new}$. The cubic scaling of the GP makes this approach less scalable, though.

Another way to define task similarity is to take the existing evaluations $P_{i,j}$, use Thompson Sampling \citep{thompson1933likelihood}
to obtain the optima distribution $\rho^{j}_{max}$, and then measure the KL-divergence \citep{kullback1951information} between $\rho^{j}_{max}$ and $\rho^{new}_{max}$ \citep{Ramachandran2018b}. These distributions are then merged into a mixture distribution based on the similarities and used to build an acquisition function that predicts the next most promising configuration to evaluate. It is so far only evaluated to tune 2 SVM hyperparameters using 5 tasks.

Finally, a complementary way to leverage $\mathbf{P}$ is to recommend which configurations should \emph{not} be used. After training surrogate models per task, we can look up which $t_{j}$ are most similar to $t_{new}$, and then use $s_{j}(\theta_{i})$ to discover regions of $\Theta$ where performance is predicted to be poor. Excluding these regions can speed up the search for better-performing ones. \citet{Wistuba2015b} do this using a task similarity measure based on the Kendall tau rank correlation coefficient \citep{kendall1938new} between the ranks obtained by ranking configurations $\theta_{i}$ using $P_{i,j}$ and $P_{i,new}$, respectively. 



\subsection{Learning Curves}
\label{sec:curves}
We can also extract meta-data about the training process itself, such as how fast model performance improves as more training data is added. If we divide the training in steps $s_{t}$,  usually adding a fixed number of training examples every step, we can measure the performance $P(\theta_{i},t_{j},s_{t})$ = $P_{i,j,t}$ of configuration $\theta_{i}$ on task $t_{j}$ after step $s_{t}$, yielding a \emph{learning curve} across the time steps $s_{t}$. Learning curves are used extensively to speed up hyperparameter optimization on a given task \citep{kohavi1995automatic,provost1999efficient,swersky2014freeze,chandrashekaran2017speeding}. In meta-learning, however, learning curve information is transferred across tasks.

While evaluating a configuration on new task $t_{new}$, we can halt the training after a certain number of iterations $r < t$, and use the partially observed learning curve to predict how well the configuration will perform on the full dataset based on prior experience with other tasks, and decide whether to continue the training or not. This can significantly speed up the search for good configurations. 

One approach is to assume that similar tasks yield similar learning curves. First, define a distance between tasks based on how similar the partial learning curves are: $dist(t_{a},t_{b}) = f(P_{i,a,t},P_{i,b,t})$ with $t=1,...,r$. Next, find the $k$ most similar tasks $t_{1..k}$ and use their complete learning curves to predict how well the configuration will perform on the new complete dataset. Task similarity can be measured by comparing the shapes of the partial curves across all configurations tried, and the prediction is made by adapting the `nearest' complete curve(s) to the new partial curve \citep{Leite:2005p725,Leite:2007p6022}. This approach was also successful in combination with active testing \citep{Leite:2010}, and can be sped up further by using multi-objective evaluation measures that include training time \citep{vanrijn2015}.

Interestingly, while several methods aim to predict learning curves during neural architecture search \citep{elsken2018neural}, as of yet none of this work leverages learning curves previously observed on other tasks.


\section{Learning from Task Properties}
\label{sec:metafeatures}
Another rich source of meta-data are characterizations (meta-features) of the task at hand. Each task $t_{j} \in T$ is described with a vector $m(t_j) = (m_{j,1},...,m_{j,K})$ of $K$ meta-features $m_{j,k} \in M$, the set of all known meta-features. This can be used to define a task similarity measure based on, for instance, the Euclidean distance between $m(t_i)$ and $m(t_j)$, so that we can transfer information from the most similar tasks to the new task $t_{new}$. Moreover, together with prior evaluations $\textbf{P}$, we can train a \emph{meta-learner} $L$ to predict the performance $P_{i,new}$ of configurations $\theta_{i}$ on a new task $t_{new}$.

\subsection{Meta-Features}
Table \ref{tab:metafeatures} provides a concise overview of the most commonly used meta-features, together with a short rationale for why they are indicative of model performance. Where possible, we also show the formulas to compute them. More complete surveys can be found in the literature \citep{rivolli2018,Vanschoren2010,Mantovani:2018,Reif2014,Castiello2005}.

\begin{table}[hp]
\footnotesize
\begin{minipage}{\textwidth}
\begin{center}
\scalebox{1}{
\begin{tabular}{llll}
  \toprule
  Name & Formula & Rationale &  Variants \\
  \midrule
   Nr instances & $n$ & Speed, Scalability \citep{Michie1994} & $p/n$, $log(n)$, log(n/p)  \\ 
   Nr features & $p$ & Curse of dimensionality \citep{Michie1994} & $log(p)$, \% categorical \\ 
   Nr classes & $c$ & Complexity, imbalance \citep{Michie1994} & ratio min/maj class \\ 
   Nr missing values & $m$ & Imputation effects \citep{kalousis02} & \% missing \\ 
   Nr outliers & $o$ & Data noisiness \citep{Rousseeuw2011} & $o/n$ \\
   \midrule
   Skewness & $\frac{E(X-\mu_{X})^{3}}{\sigma_{X}^{3}}$ & Feature normality \citep{Michie1994} & min,max,$\mu$,$\sigma$,$q_{1},q_{3}$\\
   Kurtosis & $\frac{E(X-\mu_{X})^{4}}{\sigma_{X}^{4}}$ & Feature normality \citep{Michie1994} & min,max,$\mu$,$\sigma$,$q_{1},q_{3}$\\
   Correlation & $\rho_{X_{1}X_{2}}$ & Feature interdependence \citep{Michie1994} & min,max,$\mu$,$\sigma$,$\rho_{XY}$\\
   Covariance & $cov_{X_{1}X_{2}}$ & Feature interdependence \citep{Michie1994} & min,max,$\mu$,$\sigma$,$cov_{XY}$ \\
   Concentration & $\tau_{X_{1}X_{2}}$ & Feature interdependence \citep{Kalousis2001a} & min,max,$\mu$,$\sigma$,$\tau_{XY}$\\
   Sparsity & sparsity(X) & Degree of discreteness \citep{Salama2013} & min,max,$\mu$,$\sigma$ \\
   Gravity & gravity(X) & Inter-class dispersion \citep{Ali2006} & \\  
   ANOVA p-value & $p_{val_{\texttt{X}_{1}X_{2}}}$ & Feature redundancy \citep{kalousis02} & $p_{val_{XY}}$\citep{soares+04} \\
   Coeff. of variation & $\frac{\sigma_{Y}}{\mu_{Y}}$ & Variation in target \citep{soares+04} & \\ 
   PCA $\rho_{\lambda_{1}}$ & $\sqrt{\frac{\lambda_{1}}{1+\lambda_{1}}}$ & Variance in first PC \citep{Michie1994} & $\frac{\lambda_{1}}{\sum_{i} \lambda_{i}}$\citep{Michie1994} \\
   PCA skewness & & Skewness of first PC \citep{feurer2014using} & PCA kurtosis \\
   PCA 95\% & $\frac{dim_{95\% var}}{p}$ & Intrinsic dimensionality \citep{bardenet2013collaborative} & \\
   Class probability & $P(\texttt{C})$ & Class distribution \citep{Michie1994} & min,max,$\mu$,$\sigma$\\
   \midrule
   Class entropy  & $H(\texttt{C})$ & Class imbalance \citep{Michie1994} & \\
   Norm. entropy & $\frac{H(\texttt{X})}{log_{2}n}$ & Feature informativeness \citep{Castiello2005} & min,max,$\mu$,$\sigma$ \\
   Mutual inform. & $MI(\texttt{C},\texttt{X})$ & Feature importance \citep{Michie1994} & min,max,$\mu$,$\sigma$ \\
   Uncertainty coeff. & $\frac{MI(\texttt{C},\texttt{X})}{H(\texttt{C})}$ & Feature importance \citep{Agresti:2002p7509} & min,max,$\mu$,$\sigma$ \\
   Equiv. nr. feats & $\frac{H(C)}{\overline{MI(C,X)}}$ & Intrinsic dimensionality \citep{Michie1994} & \\
   Noise-signal ratio & $\frac{\overline{H(X)}-\overline{MI(C,X)}}{\overline{MI(C,X)}}$ & Noisiness of data \citep{Michie1994} & \\
   \midrule
   Fisher's discrimin. & $\frac{(\mu_{c1}-\mu_{c2})^{2}}{\sigma_{c1}^{2}-\sigma_{c2}^{2}}$ & Separability classes $c_{1},c_{2}$ \citep{Ho:2002} & See \citet{}{Ho:2002} \\
   Volume of overlap & & Class distribution overlap \citep{Ho:2002} & See \citet{Ho:2002} \\
   Concept variation & & Task complexity \citep{Vilalta:2002p5805} & See \citet{Vilalta:1999p5745} \\
   Data consistency & & Data quality \citep{Kopf:2002p5864} & See \citet{Kopf:2002p5864} \\
   \midrule
   Nr nodes, leaves & $|\eta|,|\psi|$& Concept complexity \citep{Peng:2002p705} & Tree depth \\
   Branch length & & Concept complexity \citep{Peng:2002p705} & min,max,$\mu$,$\sigma$ \\
   Nodes per feature & $|\eta_{X}|$ & Feature importance \citep{Peng:2002p705} & min,max,$\mu$,$\sigma$ \\
   Leaves per class & $\frac{|\psi_{c}|}{|\psi|}$ &  Class complexity \citep{Filchenkov2015} & min,max,$\mu$,$\sigma$ \\
   Leaves agreement & $\frac{n_{\psi_{i}}}{n}$ &  Class separability \citep{Bensusan2000} & min,max,$\mu$,$\sigma$ \\
   Information gain & & Feature importance \citep{Bensusan2000} & min,max,$\mu$,$\sigma$, gini \\  
   \midrule
   Landmarker(1NN) & $P(\theta_{1NN},t_{j})$ & Data sparsity \citep{Pfahringer:2000p553} & See \citet{Pfahringer:2000p553}\\
   Landmarker(Tree) & $P(\theta_{Tree},t_{j})$ & Data separability \citep{Pfahringer:2000p553} & Stump,RandomTree \\
   Landmarker(Lin) & $P(\theta_{Lin},t_{j})$ & Linear separability \citep{Pfahringer:2000p553} & Lin.Disciminant\\
   Landmarker(NB) & $P(\theta_{NB},t_{j})$ & Feature independence \citep{Pfahringer:2000p553} & See \citet{Ler:2005p1680}\\
   Relative LM & $P_{a,j} - P_{b,j}$ & Probing performance \citep{Furnkranz:2001p1278} & \\
   Subsample LM & $P(\theta_{i},t_{j},s_{t})$ & Probing performance \citep{Soares:2001p708} & \\
  \bottomrule
\end{tabular}
}
\end{center}
\end{minipage}
\caption{Overview of commonly used meta-features. Groups from top to bottom: simple, statistical, information-theoretic, complexity, model-based, and landmarkers. Continuous features $X$ and target $Y$ have mean $\mu_{X}$, stdev $\sigma_{X}$, variance $\sigma^{2}_{X}$. Categorical features $\texttt{X}$ and class $\texttt{C}$ have categorical values $\pi_{i}$, conditional probabilities $\pi_{i|j}$, joint probabilities $\pi_{i,j}$, marginal probabilities $\pi_{i+}=\sum_{j}\pi_{ij}$, entropy $H($\texttt{X}$)=-\sum_{i}\pi_{i+}log_{2}(\pi_{i+})$.}
\label{tab:metafeatures}
\end{table}


To build a meta-feature vector $m(t_j)$, one needs to select and further process these meta-features. Studies on OpenML meta-data have shown that the optimal set of meta-features depends on the application \citep{Bilalli2017}. Many meta-features are computed on single features, or combinations of features, and need to be aggregated by summary statistics (min,max,$\mu$,$\sigma$,quartiles,...) or histograms \citep{Kalousis2001a}. One needs to systematically extract and aggregate them \citep{Pinto2016}. When computing task similarity, it is also important to normalize all meta-features \citep{bardenet2013collaborative}, perform feature selection \citep{Todorovski:2000p5454}, or employ dimensionality reduction techniques (e.g. PCA) \citep{Bilalli2017}. When learning meta-models, one can also use relational meta-learners \citep{Todorovski:1999p5595} or case-based reasoning methods \citep{lindner+99,Hilario:2001p001,Kalousis:2003p335}.

Beyond these general-purpose meta-features, many more specific ones were formulated. For streaming data one can use streaming landmarks \citep{vanRijn2018,van2014algorithm}, for time series data one can compute autocorrelation coefficients or the slope of regression models \citep{Arinze:1994p9871,Prudencio:2004p6308,dosSantos:2004p6355}, and for unsupervised problems one can cluster the data in different ways and extract properties of these clusters \citep{Soares:2009p9657}. In many applications, domain-specific information can be leveraged as well \citep{Smith-Miles2008,Olier2018}.

\subsection{Learning Meta-Features}
Instead of manually defining meta-features, we can also \emph{learn} a joint representation for groups of tasks. One approach is to build meta-models that generate a landmark-like meta-feature representation $M'$ given other task meta-features $M$ and trained on performance meta-data $\textbf{P}$, or $f: M \mapsto M'$. \citet{Sun:2013} do this by evaluating a predefined set of configurations $\theta_{i}$ on all prior tasks $t_{j}$, and generating a binary metafeature $m_{j,a,b} \in M'$ for every pairwise combination of configurations $\theta_{a}$ and $\theta_{b}$, indicating whether $\theta_{a}$ outperformed $\theta_{b}$ or not, thus $m'(t_j) = (m_{j,a,b},m_{j,a,c},m_{j,b,c},...)$. To compute $m_{new,a,b}$, \emph{meta-rules} are learned for every pairwise combination (a,b), each predicting whether $\theta_{a}$ will outperform $\theta_{b}$ on task $t_{j}$, given its other meta-features $m(t_j)$.

We can also learn a joint representation based entirely on the available $\textbf{P}$ meta-data, i.e. $f: \textbf{P} \times \Theta \mapsto M'$. We previously discussed how to do this with feed-forward neural nets \citep{perrone2017multiple} in Section \ref{sec:hptransfer}. If the tasks share the same input space, e.g., they are images of the same resolution, one can also use Siamese networks to learn a meta-feature representation \citep{kim2017}. These are trained by feeding the data of two different tasks to two twin networks, and using the differences between the predicted and observed performance $P_{i,new}$ as the error signal. Since the model parameters between both networks are tied in a Siamese network, two very similar tasks are mapped to the same regions in the latent meta-feature space. They can be used for warm starting Bayesian hyperparameter optimization \citep{kim2017} and neural architecture search \citep{NurAfif2018}.

\subsection{Warm-Starting Optimization from Similar Tasks}
\label{sec:warmstart}
Meta-features are a very natural way to estimate task similarity and initialize optimization procedures based on promising configurations on similar tasks. This is akin to how human experts start a manual search for good models, given experience on related tasks. 

Starting a \emph{genetic search} algorithm in regions of the search space with promising solutions can significantly speed up convergence to a good solution. Gomes et al. \citep{gomes2012combining} recommend initial configurations by finding the $k$ most similar prior tasks $t_{j}$ based on the L1 distance between vectors $m(t_j)$ and $m(t_{new})$, where each $m(t_j)$ includes 17 simple and statistical meta-features. For each of the $k$ most similar tasks, the best configuration is evaluated on $t_{new}$, and used to initialize a genetic search algorithm (Particle Swarm Optimization), as well as Tabu Search. \citet{reif2012meta} follow a very similar approach, using 15 simple, statistical, and landmarking meta-features. They use a forward selection technique to find the most useful meta-features, and warm-start a standard genetic algorithm (GAlib) with a modified Gaussian mutation operation. Variants of active testing (see Sect. \ref{sec:hptransfer}) that use meta-features were also tried \citep{Miranda:2013,leite+12}, but did not perform better than the approaches based on relative landmarks.

Also model-based optimization approaches can benefit greatly from an initial set of promising configurations. SCoT \citep{bardenet2013collaborative} trains a single surrogate ranking model $f: M \times \Theta \rightarrow R$, predicting the rank of $\theta_{i}$ on task $t_{j}$. M contains 4 meta-features (3 simple ones and one based on PCA). The surrogate model is trained on all the rankings, including those on $t_{new}$. Ranking is used because the scale of evaluation values can differ greatly between tasks. A GP regression converts the ranks to probabilities to do Bayesian optimization, and each new $P_{i,new}$ is used to retrain the surrogate model after every step.  

\citet{Schilling2015} use a modified multilayer perceptron as a surrogate model, of the form $s_{j}(\theta_{i},m(t_j),b(t_j)) = P_{i,j}$ where $m(t_j)$ are the meta-features and $b(t_j)$ is a vector of $j$ binary indications which are 1 if the meta-instance is from $t_{j}$ and 0 otherwise. The multi-layer perceptron uses a modified activation function based on factorization machines \citep{Rendle2010} in the first layer, aimed at learning a latent representation for each task to model task similarities. Since this model cannot represent uncertainties, an ensemble of 100 multilayer perceptrons is trained to get predictive means and simulate variances.

Training a single surrogate model on all prior meta-data is often less scalable. \citet{yogatama2014efficient} also build a single Bayesian surrogate model, but only include tasks similar to $t_{new}$, where task similarity is defined as the Euclidean distance between meta-feature vectors consisting of 3 simple meta-features. The $P_{i,j}$ values are standardized to overcome the problem of different scales for each $t_{j}$. The surrogate model learns a Gaussian process with a specific kernel combination on all instances.

\citet{feurer2014using} offer a simpler, more scalable method that warm-starts Bayesian optimization by sorting all prior tasks $t_{j}$ similar to \citet{gomes2012combining}, but including 46 simple, statistical, and landmarking meta-features, as well as $H(\texttt{C})$. The $t$ best configurations on the $d$ most similar tasks are used to warm-start the surrogate model. They search over many more hyperparameters than earlier work, including preprocessing steps. This warm-starting approach was also used very effectively, and combined with ensembling, in autosklearn \citep{Feurer2015}. 

Finally, one can also use \emph{collaborative filtering} to recommend promising configurations \citep{stern2010collaborative}. By analogy, the tasks $t_{j}$ (users) provide ratings ($P_{i,j}$) for the configurations $\theta_{i}$ (items), and matrix factorization techniques are used to predict unknown $P_{i,j}$ values and recommend the best configurations for any task. An important issue here is the cold start problem, since the matrix factorization requires at least some evaluations on $t_{new}$. \citet{yang2018} use a D-optimal experiment design to sample an initial set of evaluations $P_{i,new}$. They predict both the predictive performance and runtime, to recommend a set of warm-start configurations that are both accurate and fast. \citet{misir:hal-00922840} and \citet{misir2017alors} leverage meta-features to solve the cold start problem. \citet{fusi2017probabilistic} also use meta-features, following the same procedure as \citet{Feurer2015}, and use a probabilistic matrix factorization approach that allows them to perform Bayesian optimization to further optimize their pipeline configurations $\theta_{i}$. This approach also yields useful latent embeddings of both the tasks and configurations.

\subsection{Meta-Models}
We can also \emph{learn} the complex relationship between a task's meta-features and the utility of specific configurations by building a meta-model $L$ that recommends the most useful configurations $\Theta^{*}_{new}$ given the meta-features $M$ of the new task $t_{new}$. There exists a rich body of earlier work \citep{Brazdil2009,lemke+15,giraud2008metalearning,luo2016review} on building meta-models for algorithm selection \citep{Bensusan2000a,Pfahringer:2000p553,kalousis02,Bischl2016} and hyperparameter recommendation \citep{kuba+02a,soares+04,Ali2006a,Nisioti2018}. Experiments showed that boosted and bagged trees often yielded the best predictions, although much depends on the exact meta-features used \citep{Kalousis2001a,Kopf:2002p5864}. 

\subsubsection{Ranking}
Meta-models can also generate a \emph{ranking} of the top-$K$ most promising configurations. One approach is to build a k-nearest neighbor (kNN) meta-model to predict which tasks are similar, and then rank the best configurations on these similar tasks \citep{Brazdil2003,dosSantos:2004p6355}. This is similar to the work discussed in Section \ref{sec:warmstart}, but without ties to a follow-up optimization approach. Meta-models specifically meant for ranking, such as predictive clustering trees \citep{todorovski+02} and label ranking trees \citep{cheng2009decision} were also shown to work well. Approximate Ranking Trees Forests (ART Forests) \citep{Sun:2013}, ensembles of fast ranking trees, prove to be especially effective, since they have `built-in' meta-feature selection, work well even if few prior tasks are available, and the ensembling makes the method more robust. \emph{autoBagging} \citep{Pinto:2017} ranks Bagging workflows including four different Bagging hyperparameters, using an XGBoost-based ranker, trained on $140$ OpenML datasets and $146$ meta-features. \citet{Lorena:2018} recommend SVM configurations for regression problems using a kNN meta-model and a new set of meta-features based on data complexity.

\subsubsection{Performance Prediction}
Meta-models can also directly predict the performance, e.g. accuracy or training time, of a configuration on a given task, given its meta-features. This allows us to estimate whether a configuration will be interesting enough to evaluate in any optimization procedure. Early work used linear regression or rule-base regressors to predict the performance of a discrete set of configurations and then rank them accordingly \citep{Bensusan:2001p344,kopf+00}. Guerra et al. \citep{guerra2008predicting} train an SVM meta-regressor per classification algorithm to predict its accuracy, under default settings, on a new task $t_{new}$ given its meta-features. Reif et al. \citep{Reif2014} train a similar meta-regressor on more meta-data to predict its \emph{optimized} performance. Davis et al. \citep{Davis2018} use a MultiLayer Perceptron based meta-learner instead, predicting the performance of a specific algorithm configuration. 

Instead of predicting predictive performance, a meta-regressor can also be trained to predict algorithm training/prediction time, for instance, using an SVM regressor trained on meta-features \citep{Reif2011}, itself tuned via genetic algorithms~\citep{Priya:2012}. \citet{yang2018} predict configuration runtime using polynomial regression, based only on the number of instances and features. \citet{Hutter+2014} provide a general treatise on predicting algorithm runtime in various domains.

Most of these meta-models generate promising configurations, but don't actually tune these configurations to $t_{new}$ themselves. Instead, the predictions can be used to warm-start or guide any other optimization technique, which allows for all kinds of combinations of meta-models and optimization techniques. Indeed, some of the work discussed in Section \ref{sec:warmstart} can be seen as using a distance-based meta-model to warm-start Bayesian optimization \citep{feurer2014using,fusi2017probabilistic} or evolutionary algorithms \citep{gomes2012combining,reif2012meta}. In principle, other meta-models could be used here as well.

Instead of learning the relationship between a task's meta-features and configuration performance, one can also build surrogate models predicting the performance of configurations on specific tasks\citep{Eggensperger+18}. One can then learn how to combine these per-task predictions to warm-start or guide optimization techniques on a new task $t_{new}$ \citep{Feurer2018,perrone2017multiple,Springenberg2016,Wistuba2018}, as discussed in Section \ref{sec:hptransfer}. While meta-features could also be used to combine per-task predictions based on task similarity, it is ultimately more effective to gather new observations $P_{i,new}$, since these allow to refine the task similarity estimates with every new observation \citep{feurerscalable,Wistuba2018,leite+12}.





\subsection{Pipeline Synthesis}
When creating entire machine learning pipelines \citep{Serban2013}, the number of configuration options grows dramatically, making it even more important to leverage prior experience. One can control the search space by imposing a fixed structure on the pipeline, fully described by a set of hyperparameters. One can then use the most promising pipelines on similar tasks to warm-start a Bayesian optimization \citep{Feurer2015,fusi2017probabilistic}. 

Other approaches give recommendations for certain pipeline steps \citep{Post2016,Strang2018}, and can be leveraged in larger pipeline construction approaches, such as planning \citep{nguyen2014using,kietz2012designing,gilp4ml,weverml} or evolutionary techniques \citep{olson2016evaluation,Sun2013a}. \citet{nguyen2014using} construct new pipelines using a beam search focussed on components recommended by a meta-learner, and is itself trained on examples of successful prior pipelines. \citet{Bilalli2018} predict which pre-processing techniques are recommended for a given classification algorithm. They build a meta-model per target classification algorithm that, given the $t_{new}$ meta-features, predicts which preprocessing technique should be included in the pipeline. Similarly, \citet{Schoenfeld2018} build meta-models predicting when a preprocessing algorithm will improve a particular classifier's accuracy or runtime. 

AlphaD3M \citep{drorialphad3m} uses a \emph{self-play} reinforcement learning approach in which the current state is represented by the current pipeline, and actions include the addition, deletion, or replacement of pipeline components. A Monte Carlo Tree Search (MCTS) generates pipelines, which are evaluated to train a recurrent neural network (LSTM) that can predict pipeline performance, in turn producing the action probabilities for the MCTS in the next round. The state description also includes meta-features of the current task, allowing the neural network to learn across tasks.

\subsection{To Tune or Not to Tune?}
To reduce the number of configuration parameters to be optimized, and to save valuable optimization time in time-constrained settings, meta-models have also been proposed to predict whether or not it is worth tuning a given algorithm \emph{given the meta-features of the task at hand} \citep{Ridd:2014} and how much improvement we can expect from tuning a specific algorithm versus the additional time investment \citep{Sanders8215600}. More focused studies on specific learning algorithms yielded meta-models predicting when it is necessary to tune SVMs \citep{mantovani2015tune}, what are good default hyperparameters for SVMs given the task (including interpretable meta-models) \citep{mantovani2015meta}, and how to tune decision trees \citep{mantovani2016hyper}.



%
%

\section{Learning from Prior Models}
\label{sec:modeltransfer}
The final type of meta-data we can learn from are prior machine learning models themselves, i.e., their structure and learned model parameters. In short, we want to train a \emph{meta-learner} $L$ that learns how to train a (base-) learner $l_{new}$ for a new task $t_{new}$, given similar tasks $t_{j} \in T$ and the corresponding optimized models $l_{j} \in \mathcal{L}$, where $\mathcal{L}$ is the space of all possible models. 
The learner $l_{j}$ is typically defined by its model parameters $W = \{w_{k}\}$, $k = 1 .. K$  and/or its configuration $\theta_{i} \in \Theta$.


\subsection{Transfer Learning}
In \emph{transfer learning} \citep{Thrun98A}, we take models trained on one or more \emph{source} tasks $t_{j}$, and use them as starting points for creating a model on a similar \emph{target} task $t_{new}$. This can be done by forcing the target model to be structurally or otherwise similar to the source model(s). This is a generally applicable idea, and transfer learning approaches have been proposed for kernel methods \citep{Evgeniou05,Evgeniou04}, parametric Bayesian models \citep{Rosenstein05,Raina05,Bakker03}, Bayesian networks \citep{Niculescu05}, clustering \citep{Thrun98} and reinforcement learning \citep{Hengst02,Dietterich02}. Neural networks, however, are exceptionally suitable for transfer learning because both the structure and the model parameters of the source models can be used as a good initialization for the target model, yielding a \emph{pre-trained} model which can then be further fine-tuned using the available training data on $t_{new}$ \citep{Thrun95A,Baxter96,Bengio2012,caruana1995learning}. In some cases, the source network may need to be modified before transferring it \citep{Sharkey93}. We will focus on neural networks in the remainder of this section. 

Especially large image datasets, such as ImageNet \citep{krizhevsky2012imagenet}, have been shown to yield pre-trained models that transfer exceptionally well to other tasks \citep{donahue2014decaf,sharif2014cnn}. However, it has also been shown that this approach doesn't work well when the target task is not so similar \citep{yosinski2014transferable}. Rather than hoping that a pre-trained model `accidentally' transfers well to a new problem, we can purposefully imbue meta-learners with an inductive bias (learned from many similar tasks) that allows them to learn new tasks much faster, as we will discuss below.


\subsection{Meta-Learning in Neural Networks}
An early meta-learning approach is to create recurrent neural networks (RNNs) able to modify their own weights \citep{schmidhuber1992learning,schmidhuber1993neural}. During training, they use their own weights as additional input data and observe their own errors to learn how to modify these weights in response to the new task at hand. The updating of the weights is defined in a parametric form that is differentiable end-to-end and can jointly optimize both the network and training algorithm using gradient descent, yet is also very difficult to train. Later work used reinforcement learning across tasks to adapt the search strategy \citep{schmidhuber1997shifting} or the learning rate for gradient descent \citep{daniel2016learning} to the task at hand.

Inspired by the feeling that backpropagation is an unlikely learning mechanism for our own brains, \citet{bengio1995search} replace backpropagation with simple biologically-inspired parametric rules (or evolved rules \citep{chalmers1991evolution}) to update the synaptic weights. The parameters are optimized, e.g. using gradient descent or evolution, across a set of input tasks. \citet{runarsson2000evolution} replaced these parametric rules with a single layer neural network. \citet{santoro2016one} instead use a memory-augmented neural network to learn how to store and retrieve `memories' of prior classification tasks. \citet{hochreiter+01} use LSTMs \citep{hochreiter1997long} as a meta-learner to train multi-layer perceptrons.

\citet{andrychowicz2016learning} also replace the optimizer, e.g. stochastic gradient descent, with an LSTM trained on multiple prior tasks. The loss of the meta-learner (optimizer) is defined as the sum of the losses of the base-learners (optimizees), and optimized using gradient descent. At every step, the meta-learner chooses the weight update estimated to reduce the optimizee's loss the most, based on the learned model weights $\{w_{k}\}$ of the previous step as well as the current performance gradient. Later work generalizes this approach by training an optimizer on synthetic functions, using gradient descent \citep{chen2016learning}. This allows meta-learners to optimize optimizees even if these do not have access to gradients.

In parallel, \citet{li2016learning} proposed a framework for learning optimization algorithms from a reinforcement learning perspective. It represents any particular optimization algorithm as a policy, and then learns this policy via guided policy search. Follow-up work \citep{li2017learning} shows how to leverage this approach to learn optimization algorithms for (shallow) neural networks.

The field of \emph{neural architecture search} includes many other methods that build a model of neural network performance for a specific task, for instance using Bayesian optimization or reinforcement learning. See \citet{elsken2018neural} for an in-depth discussion. However, most of these methods do not (yet) generalize across tasks and are therefore not discussed here.  



\subsection{Few-Shot Learning}
A particularly challenging meta-learning problem is to train an accurate deep learning model using only a few training examples, given prior experience with very similar tasks for which we have large training sets available. This is called \emph{few-shot learning}. Humans have an innate ability to do this, and we wish to build machine learning agents that can do the same \citep{lake2017building}.  A particular example of this is `K-shot N-way' classification, in which we are given many examples (e.g., images) of certain classes (e.g., objects), and want to learn a classifier $l_{new}$ able to classify $N$ new classes using only $K$ examples of each. 

Using prior experience, we can, for instance, learn a common feature representation of all the tasks, start training $l_{new}$ with a better model parameter initialization $W_{init}$ and acquire an inductive bias that helps guide the optimization of the model parameters, so that $l_{new}$ can be trained much faster than otherwise possible.

Earlier work on \emph{one-shot} learning is largely based on hand-engineered features \citep{fei2006one,fei2006knowledge,fink2005object,bart2005cross}. With meta-learning, however, we hope to learn a common feature representation for all tasks in an end-to-end fashion. 

\citet{vinyals2016matching} state that, to learn from very little data, one should look to non-parameteric models (such as k-nearest neighbors), which use a memory component rather than learning many model parameters. Their meta-learner is a Matching Network that apply the idea of a memory component in a neural net. It learns a common representation for the labelled examples, and \emph{matches} each new test instance to the memorized examples using cosine similarity. The network is trained on minibatches with only a few examples of a specific task each.

\citet{snell2017prototypical} propose Prototypical Networks, which map examples to a p-dimensional vector space such that examples of a given output class are close together. It then calculates a prototype (mean vector) for every class. New test instances are mapped to the same vector space and a distance metric is used to create a softmax over all possible classes. \citet{ren2018meta} extend this approach to semi-supervised learning.

\citet{ravi2016optimization} use an LSTM-based meta-learner to learn an update rule for training a neural network learner. With every new example, the learner returns the current gradient and loss to the LSTM meta-learner, which then updates the model parameters $\{w_{k}\}$ of the learner. The meta-learner is trained across all prior tasks.

Model-Agnostic Meta-Learning (MAML) \citep{pmlr-v70-finn17a}, on the other hand, does not try to learn an update rule, but instead learns a model parameter initialization $W_{init}$ that generalizes better to similar tasks. 
Starting from a random $\{w_{k}\}$, it iteratively selects a batch of prior tasks, and for each it trains the learner on $K$ examples to compute the gradient and loss (on a test set). It then backpropagates the \emph{meta-gradient} to update the weights $\{w_{k}\}$ in the direction in which they would have been easier to update. In other words, after each iteration, the weights $\{w_{k}\}$ become a better $W_{init}$ to start finetuning any of the tasks. \citet{finn2017meta} show that MAML is able to approximate any learning algorithm when using a sufficiently deep ReLU network and certain losses. They also conclude that the MAML initializations are more resilient to overfitting on small samples, and generalize more widely than meta-learning approaches based on LSTMs. \citet{grant2018recasting} present a novel derivation of and extension to MAML, illustrating that this
algorithm can be understood as inference for the parameters of a prior distribution in a hierarchical
Bayesian model.

REPTILE \citep{Nichol2018} is an approximation of MAML that executes stochastic gradient descent for $K$ iterations on a given task, and then gradually moves the initialization weights in the direction of the weights obtained after the $K$ iterations. The intuition is that every task likely has more than one set of optimal weights $\{w^{*}_{i}\}$, and the goal is to find a $W_{init}$ that is close to at least one of those $\{w^{*}_{i}\}$ for every task.

Finally, we can also derive a meta-learner from a black-box neural network. 
\citet{santoro2016meta} propose Memory-Augmented Neural Networks (MANNs), which train a Neural Turing Machine (NTM) \citep{graves2014neural}, a neural network with augmented memory capabilities, as a meta-learner. This meta-learner can then memorize information about previous tasks and leverage that to learn a learner $l_{new}$.
SNAIL \citep{mishra2018simple} is a generic meta-learner architecture consisting of interleaved temporal convolution and causal attention layers. The convolutional networks learn a common feature vector for the training instances (images) to aggregate information from past experiences. The causal attention layers learn which pieces of information to pick out from the gathered experience to generalize to new tasks.

Overall, the intersection of deep learning and meta-learning proves to be particular fertile ground for groundbreaking new ideas, and we expect this field to become more important over time.

\subsection{Beyond Supervised Learning}
Meta-learning is certainly not limited to (semi-)supervised tasks, and has been successfully applied to solve tasks as varied as reinforcement learning, active learning, density estimation and item recommendation. The base-learner may be unsupervised while the meta-learner is supervised, but other combinations are certainly possible as well.

\citet{duan2016rl} propose an end-to-end reinforcement learning (RL) approach consisting of a task-specific \emph{fast} RL algorithm which is guided by a general-purpose \emph{slow} meta-RL algorithm. The tasks are interrelated Markov Decision Processes (MDPs). The meta-RL algorithm is modeled as an RNN, which receives the observations, actions, rewards and termination flags. The activations of the RNN store the state of the fast RL learner, and the RNN's weights are learned by observing the performance of fast learners across tasks.  

In parallel, \citet{wang2016learning} also proposed to use a deep RL algorithm to train an RNN, receiving the actions and rewards of the previous interval in order to learn a base-level RL algorithm for specific tasks. Rather than using relatively unstructured tasks such as random MDPs, they focus on structured task distributions (e.g., dependent bandits) in which the meta-RL algorithm can exploit the inherent task structure.

\citet{Pang2018} offer a meta-learning approach to active learning (AL). The base-learner can be any binary classifier, and the meta-learner is a deep RL network consisting of a deep neural network that learns a representation of the AL problem across tasks, and a policy network that learns the optimal policy, parameterized as weights in the network. The meta-learner receives the current state (the unlabeled point set and base classifier state) and reward (the performance of the base classifier), and emits a query probability, i.e. which points in the unlabeled set to query next.

\citet{reed2017few} propose a few-shot approach for density estimation (DE). The goal is to learn a probability distribution over a small number of images of a certain concept (e.g., a handwritten letter) that can be used to generate images of that concept, or compute the probability that an image shows that concept. The approach uses autoregressive image models which factorize the joint distribution into per-pixel factors, usually conditioned on (many) examples of the target concept. Instead, a MAML-based few-shot learner is used, trained on examples of many other (similar) concepts.

Finally, \citet{vartak2017meta} address the cold-start problem in matrix factorization. They propose a deep neural network architecture that learns a (base) neural network whose biases are adjusted based on task information. While the structure and weights of the neural net recommenders remain fixed, the meta-learner learns how to adjust the biases based on each user's item history.

All these recent new developments illustrate that it is often fruitful to look at problems through a \emph{meta-learning lens} and find new, data-driven approaches to replace hand-engineered base-learners.


    
\section{Conclusion}
Meta-learning opportunities present themselves in many different ways, and can be embraced using a wide spectrum of learning techniques. Every time we try to learn a certain task, whether successful or not, we gain useful experience that we can leverage to learn new tasks. We should never have to start entirely from scratch. Instead, we should systematically collect our `learning exhaust' and learn from it to build AutoML systems that continuously improve over time, helping us tackle new learning problems ever more efficiently. The more new tasks we encounter, and the more similar those new tasks are, the more we can tap into prior experience, to the point that most of the required learning has already been done beforehand. The ability of computer systems to store virtually infinite amounts of prior learning experiences (in the form of meta-data) opens up a wide range of opportunities to use that experience in completely new ways, and we are only starting to learn how to learn from prior experience effectively. Yet, this is a worthy goal: learning how to learn any task empowers us far beyond knowing how to learn specific tasks.

\subsection*{Acknowledgments}
The author would like to thank Pavel Brazdil, Matthias Feurer, Frank Hutter, Raghu Rajan, and Jan van Rijn for many invaluable discussions and feedback on the manuscript. 

\newpage

\bibliography{metalearningreferences}

\begin{thebibliography}{196}
\providecommand{\natexlab}[1]{#1}
\providecommand{\url}[1]{\texttt{#1}}
\expandafter\ifx\csname urlstyle\endcsname\relax
  \providecommand{\doi}[1]{doi: #1}\else
  \providecommand{\doi}{doi: \begingroup \urlstyle{rm}\Url}\fi

\bibitem[Abdulrahman et~al.(2018)Abdulrahman, Brazdil, van Rijn, and
  Vanschoren]{Abdulrahman+18}
S.~Abdulrahman, P.~Brazdil, J.~van Rijn, and J.~Vanschoren.
\newblock {Speeding up Algorithm Selection using Average Ranking and Active
  Testing by Introducing Runtime}.
\newblock \emph{Machine Learning}, 107:\penalty0 79--108, 2018.

\bibitem[Afif(2018)]{NurAfif2018}
I.~Nur Afif.
\newblock Warm-starting deep learning model construction using meta-learning.
\newblock Master's thesis, TU Eindhoven, 2018.

\bibitem[Agresti(2002)]{Agresti:2002p7509}
A.~Agresti.
\newblock \emph{Categorical Data Analysis}.
\newblock Wiley Interscience, 2002.

\bibitem[Ali and Smith-Miles(2006{\natexlab{a}})]{Ali2006}
Shawkat Ali and Kate~A. Smith-Miles.
\newblock On learning algorithm selection for classification.
\newblock \emph{Applied Soft Computing}, 6\penalty0 (2):\penalty0 119 -- 138,
  2006{\natexlab{a}}.

\bibitem[Ali and Smith-Miles(2006{\natexlab{b}})]{Ali2006a}
Shawkat Ali and Kate~A. Smith-Miles.
\newblock Metalearning approach to automatic kernel selection for support
  vector machines.
\newblock \emph{Neurocomput.}, 70\penalty0 (1):\penalty0 173--186,
  2006{\natexlab{b}}.

\bibitem[Andrychowicz et~al.(2016)Andrychowicz, Denil, Gomez, Hoffman, Pfau,
  Schaul, Shillingford, and De~Freitas]{andrychowicz2016learning}
Marcin Andrychowicz, Misha Denil, Sergio Gomez, Matthew~W Hoffman, David Pfau,
  Tom Schaul, Brendan Shillingford, and Nando De~Freitas.
\newblock Learning to learn by gradient descent by gradient descent.
\newblock In \emph{Advances in Neural Information Processing Systems}, pages
  3981--3989, 2016.

\bibitem[Arinze(1994)]{Arinze:1994p9871}
B~Arinze.
\newblock Selecting appropriate forecasting models using rule induction.
\newblock \emph{Omega}, 22\penalty0 (6):\penalty0 647--658, 1994.

\bibitem[Bakker and Heskes(2003)]{Bakker03}
B.~Bakker and T.~Heskes.
\newblock {Task Clustering and Gating for {B}ayesian Multitask Learning}.
\newblock \emph{Journal of Machine Learning Research}, 4:\penalty0 83--999,
  2003.

\bibitem[Bardenet et~al.(2013)Bardenet, Brendel, K{\'e}gl, and
  Sebag]{bardenet2013collaborative}
R{\'e}mi Bardenet, M{\'a}ty{\'a}s Brendel, Bal{\'a}zs K{\'e}gl, and Michele
  Sebag.
\newblock Collaborative hyperparameter tuning.
\newblock In \emph{Proceedings of ICML 2013}, pages 199--207, 2013.

\bibitem[Bart and Ullman(2005)]{bart2005cross}
Evgeniy Bart and Shimon Ullman.
\newblock Cross-generalization: Learning novel classes from a single example by
  feature replacement.
\newblock In \emph{CVPR}, pages 672--679, 2005.

\bibitem[Baxter(1996)]{Baxter96}
J.~Baxter.
\newblock {Learning Internal Representations}.
\newblock In \emph{Advances in Neural Information Processing Systems, NIPS},
  1996.

\bibitem[Bengio et~al.(1995)Bengio, Bengio, and Cloutier]{bengio1995search}
Samy Bengio, Yoshua Bengio, and Jocelyn Cloutier.
\newblock On the search for new learning rules for anns.
\newblock \emph{Neural Processing Letters}, 2\penalty0 (4):\penalty0 26--30,
  1995.

\bibitem[Bengio(2012)]{Bengio2012}
Y.~Bengio.
\newblock Deep learning of representations for unsupervised and transfer
  learning.
\newblock In \emph{ICML Unsupervised and Transfer Learning}, pages 17--36,
  2012.

\bibitem[Bensusan and Kalousis(2001)]{Bensusan:2001p344}
H~Bensusan and A~Kalousis.
\newblock Estimating the predictive accuracy of a classifier.
\newblock \emph{Lecture Notes in Computer Science}, 2167:\penalty0 25--36,
  2001.

\bibitem[Bensusan and Giraud-Carrier(2000)]{Bensusan2000a}
Hilan Bensusan and Christophe Giraud-Carrier.
\newblock Discovering task neighbourhoods through landmark learning
  performances.
\newblock In \emph{PKDD}, pages 325--330, 2000.

\bibitem[Bensusan et~al.(2000)Bensusan, Giraud-Carrier, and
  Kennedy]{Bensusan2000}
Hilan Bensusan, Christophe Giraud-Carrier, and Claire Kennedy.
\newblock A higher-order approach to meta-learning.
\newblock In \emph{ILP}, pages 33 -- 42, 2000.

\bibitem[Bilalli et~al.(2017)Bilalli, Abell{\'{o}}, and
  Aluja-Banet]{Bilalli2017}
Besim Bilalli, Alberto Abell{\'{o}}, and Tom{\`{a}}s Aluja-Banet.
\newblock On the predictive power of meta-features in {OpenML}.
\newblock \emph{International Journal of Applied Mathematics and Computer
  Science}, 27\penalty0 (4):\penalty0 697 -- 712, 2017.

\bibitem[Bilalli et~al.(2018)Bilalli, Abell{\'{o}}, Aluja-Banet, and
  Wrembel]{Bilalli2018}
Besim Bilalli, Alberto Abell{\'{o}}, Tom{\`{a}}s Aluja-Banet, and Robert
  Wrembel.
\newblock Intelligent assistance for data pre-processing.
\newblock \emph{Computer Standards \& Interf.}, 57:\penalty0 101 -- 109, 2018.

\bibitem[Bischl et~al.(2016)Bischl, Kerschke, Kotthoff, Lindauer, Malitsky,
  Fr{\'e}chette, Hoos, Hutter, Leyton-Brown, Tierney, and
  Vanschoren]{Bischl2016}
B.~Bischl, P.~Kerschke, L.~Kotthoff, M.~Lindauer, Y.~Malitsky,
  A.~Fr{\'e}chette, H.~Hoos, F.~Hutter, K.~Leyton-Brown, K.~Tierney, and
  J.~Vanschoren.
\newblock {ASLib}: A benchmark library for algorithm selection.
\newblock \emph{Artificial Intelligence}, 237:\penalty0 41--58, 2016.

\bibitem[Bishop(2006)]{bishop2006pattern}
Christopher~M Bishop.
\newblock Pattern recognition and machine learning.
\newblock \emph{Springer}, 2006.

\bibitem[Brazdil et~al.(2003{\natexlab{a}})Brazdil, Soares, and
  da~Costa]{brazdil+03}
P.~Brazdil, C.~Soares, and J.~Pinto da~Costa.
\newblock Ranking learning algorithms: Using {IBL} and meta-learning on
  accuracy and time results.
\newblock \emph{Machine Learning}, 50\penalty0 (3):\penalty0 251--277,
  2003{\natexlab{a}}.

\bibitem[Brazdil et~al.(2009)Brazdil, Giraud-Carrier, Soares, and
  Vilalta]{Brazdil2009}
Pavel Brazdil, Christophe Giraud-Carrier, Carlos Soares, and Ricardo Vilalta.
\newblock \emph{Metalearning: Applications to Data Mining}.
\newblock Springer-Verlag Berlin Heidelberg, 2009.

\bibitem[Brazdil et~al.(2003{\natexlab{b}})Brazdil, Soares, and {Da
  Coasta}]{Brazdil2003}
Pavel~B. Brazdil, Carlos Soares, and Joaquim~Pinto {Da Coasta}.
\newblock {Ranking learning algorithms: Using IBL and meta-learning on accuracy
  and time results}.
\newblock \emph{Machine Learning}, 50\penalty0 (3):\penalty0 251--277,
  2003{\natexlab{b}}.

\bibitem[Caruana(1995)]{caruana1995learning}
R.~Caruana.
\newblock Learning many related tasks at the same time with backpropagation.
\newblock \emph{Neural Information Processing Systems}, pages 657--664, 1995.

\bibitem[Caruana(1997)]{Caruana97}
R.~Caruana.
\newblock {Multitask Learning}.
\newblock \emph{{Machine Learning}}, 28\penalty0 (1):\penalty0 41--75, 1997.

\bibitem[Castiello et~al.(2005)Castiello, Castellano, and
  Fanelli]{Castiello2005}
Ciro Castiello, Giovanna Castellano, and Anna~Maria Fanelli.
\newblock Meta-data: {C}haracterization of input features for meta-learning.
\newblock In \emph{2nd International Conference on Modeling Decisions for
  Artificial Intelligence (MDAI)}, pages 457 -- 468, 2005.

\bibitem[Chalmers(1991)]{chalmers1991evolution}
David~J Chalmers.
\newblock The evolution of learning: An experiment in genetic connectionism.
\newblock In \emph{Connectionist Models}, pages 81--90. Elsevier, 1991.

\bibitem[Chandrashekaran and Lane(2017)]{chandrashekaran2017speeding}
Akshay Chandrashekaran and Ian~R Lane.
\newblock Speeding up hyper-parameter optimization by extrapolation of learning
  curves using previous builds.
\newblock In \emph{Joint European Conference on Machine Learning and Knowledge
  Discovery in Databases}, pages 477--492, 2017.

\bibitem[Chen et~al.(2016)Chen, Hoffman, Colmenarejo, Denil, Lillicrap,
  Botvinick, and de~Freitas]{chen2016learning}
Yutian Chen, Matthew~W Hoffman, Sergio~G{\'o}mez Colmenarejo, Misha Denil,
  Timothy~P Lillicrap, Matt Botvinick, and Nando de~Freitas.
\newblock Learning to learn without gradient descent by gradient descent.
\newblock \emph{arXiv preprint arXiv:1611.03824}, 2016.

\bibitem[Cheng et~al.(2009)Cheng, H{\"u}hn, and
  H{\"u}llermeier]{cheng2009decision}
Weiwei Cheng, Jens H{\"u}hn, and Eyke H{\"u}llermeier.
\newblock Decision tree and instance-based learning for label ranking.
\newblock In \emph{ICML}, pages 161--168, 2009.

\bibitem[Cook et~al.(1996)Cook, Kress, and Seiford]{cook+96}
W.~D. Cook, M.~Kress, and L.~W. Seiford.
\newblock A general framework for distance-based consensus in ordinal ranking
  models.
\newblock \emph{European Journal of Operational Research}, 96\penalty0
  (2):\penalty0 392--397, 1996.

\bibitem[Daniel et~al.(2016)Daniel, Taylor, and Nowozin]{daniel2016learning}
Christian Daniel, Jonathan Taylor, and Sebastian Nowozin.
\newblock Learning step size controllers for robust neural network training.
\newblock In \emph{AAAI}, pages 1519--1525, 2016.

\bibitem[Davis and Giraud-Carrier(2018)]{Davis2018}
C.~Davis and C.~Giraud-Carrier.
\newblock Annotative experts for hyperparameter selection.
\newblock In \emph{AutoML Workshop at ICML 2018}, 2018.

\bibitem[De~Sa et~al.(2017)De~Sa, Pinto, Oliveira, and Pappa]{RECIPE}
Alex De~Sa, Walter Pinto, Luiz~Otavio Oliveira, and Gisele Pappa.
\newblock {RECIPE}: A grammar-based framework for automatically evolving
  classification pipelines.
\newblock In \emph{European Conference on Genetic Programming}, pages 246--261,
  2017.

\bibitem[Dem{\v{s}}ar(2006)]{demsar06}
J.~Dem{\v{s}}ar.
\newblock {Statistical Comparisons of Classifiers over Multiple Data Sets}.
\newblock \emph{Journal of Machine Learning Research}, 7:\penalty0 1--30, 2006.

\bibitem[Dietterich(2000)]{Dietterich:2000p4081}
T~Dietterich.
\newblock Ensemble methods in machine learning.
\newblock In \emph{International workshop on multiple classifier systems},
  pages 1--15, 2000.

\bibitem[Dietterich et~al.(2002)Dietterich, Busquets, Lopez~de Mantaras, and
  Sierra]{Dietterich02}
T.~Dietterich, D.~Busquets, R.~Lopez~de Mantaras, and C.~Sierra.
\newblock {Action Refinement in Reinforcement Learning by Probability
  Smoothing}.
\newblock In \emph{19th International Conference on Machine Learning}, pages
  107--114, 2002.

\bibitem[Donahue et~al.(2014)Donahue, Jia, Vinyals, Hoffman, Zhang, Tzeng, and
  Darrell]{donahue2014decaf}
Jeff Donahue, Yangqing Jia, Oriol Vinyals, Judy Hoffman, Ning Zhang, Eric
  Tzeng, and Trevor Darrell.
\newblock Decaf: A deep convolutional activation feature for generic visual
  recognition.
\newblock In \emph{ICML}, pages 647--655, 2014.

\bibitem[dos Santos et~al.(2004)dos Santos, Ludermir, and
  Prud{\^e}ncio]{dosSantos:2004p6355}
P~dos Santos, T~Ludermir, and R~Prud{\^e}ncio.
\newblock Selection of time series forecasting models based on performance
  information.
\newblock \emph{4th International Conference on Hybrid Intelligent Systems},
  pages 366--371, 2004.

\bibitem[Drori et~al.(2018)Drori, Krishnamurthy, Rampin, de~Paula~Lourenco,
  Ono, Cho, Silva, and Freire]{drorialphad3m}
Iddo Drori, Yamuna Krishnamurthy, Remi Rampin, Raoni de~Paula~Lourenco,
  Jorge~Piazentin Ono, Kyunghyun Cho, Claudio Silva, and Juliana Freire.
\newblock {AlphaD3M}: Machine learning pipeline synthesis.
\newblock In \emph{AutoML Workshop at ICML}, 2018.

\bibitem[Duan et~al.(2016)Duan, Schulman, Chen, Bartlett, Sutskever, and
  Abbeel]{duan2016rl}
Yan Duan, John Schulman, Xi~Chen, Peter~L Bartlett, Ilya Sutskever, and Pieter
  Abbeel.
\newblock {RL}$^2$: Fast reinforcement learning via slow reinforcement
  learning.
\newblock \emph{arXiv preprint arXiv:1611.02779}, 2016.

\bibitem[Eggensperger et~al.(2018)Eggensperger, Lindauer, Hoos, Hutter, and
  Leyton-Brown]{Eggensperger+18}
K.~Eggensperger, M.~Lindauer, H.H. Hoos, F.~Hutter, and K.~Leyton-Brown.
\newblock { Efficient Benchmarking of Algorithm Configuration Procedures via
  Model-Based Surrogates }.
\newblock \emph{Machine Learning}, 107:\penalty0 15--41, 2018.

\bibitem[Elsken et~al.(2018)Elsken, Metzen, and Hutter]{elsken2018neural}
Thomas Elsken, Jan~Hendrik Metzen, and Frank Hutter.
\newblock Neural architecture search: A survey.
\newblock \emph{arXiv preprint arXiv:1808.05377}, 2018.

\bibitem[Evgeniou and Pontil(2004)]{Evgeniou04}
T.~Evgeniou and M.~Pontil.
\newblock Regularized multi-task learning.
\newblock In \emph{Tenth Conference on Knowledge Discovery and Data Mining},
  2004.

\bibitem[Evgeniou et~al.(2005)Evgeniou, Micchelli, and Pontil]{Evgeniou05}
T.~Evgeniou, C.~Micchelli, and M.~Pontil.
\newblock {Learning Multiple Tasks with Kernel Methods}.
\newblock \emph{Journal of Machine Learning Research}, 6:\penalty0 615--637,
  2005.

\bibitem[Fei-Fei(2006)]{fei2006knowledge}
Li~Fei-Fei.
\newblock Knowledge transfer in learning to recognize visual objects classes.
\newblock In \emph{Intern. Conf. on Development and Learning}, page Art. 51,
  2006.

\bibitem[Fei-Fei et~al.(2006)Fei-Fei, Fergus, and Perona]{fei2006one}
Li~Fei-Fei, Rob Fergus, and Pietro Perona.
\newblock One-shot learning of object categories.
\newblock \emph{Pattern analysis and machine intelligence}, 28\penalty0
  (4):\penalty0 594--611, 2006.

\bibitem[Feurer et~al.(2018{\natexlab{a}})Feurer, Letham, and
  Bakshy]{Feurer2018}
M~Feurer, B~Letham, and E~Bakshy.
\newblock Scalable meta-learning for {B}ayesian optimization.
\newblock \emph{arXiv}, 1802.02219, 2018{\natexlab{a}}.

\bibitem[Feurer et~al.(2014)Feurer, Springenberg, and Hutter]{feurer2014using}
Matthias Feurer, Jost~Tobias Springenberg, and Frank Hutter.
\newblock Using meta-learning to initialize {B}ayesian optimization of
  hypxerparameters.
\newblock In \emph{International Conference on Meta-learning and Algorithm
  Selection}, pages 3 -- 10, 2014.

\bibitem[Feurer et~al.(2015)Feurer, Klein, Eggensperger, Springenberg, Blum,
  and Hutter]{Feurer2015}
Matthias Feurer, Aaron Klein, Katharina Eggensperger, Jost Springenberg, Manuel
  Blum, and Frank Hutter.
\newblock Efficient and robust automated machine learning.
\newblock In \emph{Advances in Neural Information Processing Systems 28}, pages
  2944--2952, 2015.

\bibitem[Feurer et~al.(2018{\natexlab{b}})Feurer, Letham, and
  Bakshy]{feurerscalable}
Matthias Feurer, Benjamin Letham, and Eytan Bakshy.
\newblock Scalable meta-learning for bayesian optimization using
  ranking-weighted gaussian process ensembles.
\newblock In \emph{AutoML Workshop at ICML 2018}, 2018{\natexlab{b}}.

\bibitem[Filchenkov and Pendryak(2015)]{Filchenkov2015}
Andray Filchenkov and Arseniy Pendryak.
\newblock Dataset metafeature description for recommending feature selection.
\newblock In \emph{ISMW FRUCT}, pages 11--18, 2015.

\bibitem[Fink(2005)]{fink2005object}
Michael Fink.
\newblock Object classification from a single example utilizing class relevance
  metrics.
\newblock In \emph{Neural information processing syst.}, pages 449--456, 2005.

\bibitem[Finn and Levine(2017)]{finn2017meta}
Chelsea Finn and Sergey Levine.
\newblock Meta-learning and universality.
\newblock \emph{arXiv 1710.11622}, 2017.

\bibitem[Finn et~al.(2017)Finn, Abbeel, and Levine]{pmlr-v70-finn17a}
Chelsea Finn, Pieter Abbeel, and Sergey Levine.
\newblock Model-agnostic meta-learning for fast adaptation of deep networks.
\newblock In \emph{ICML}, pages 1126--1135, 2017.

\bibitem[F{\"u}rnkranz and Petrak(2001)]{Furnkranz:2001p1278}
J~F{\"u}rnkranz and J~Petrak.
\newblock An evaluation of landmarking variants.
\newblock \emph{ECML/PKDD 2001 Workshop on Integrating Aspects of Data Mining,
  Decision Support and Meta-Learning}, pages 57--68, 2001.

\bibitem[Fusi et~al.(2017)Fusi, Sheth, and Elibol]{fusi2017probabilistic}
Nicolo Fusi, Rishit Sheth, and Huseyn~Melih Elibol.
\newblock Probabilistic matrix factorization for automated machine learning.
\newblock \emph{arXiv preprint arXiv:1705.05355}, 2017.

\bibitem[Gil et~al.(2018)Gil, Yao, Ratnakar, Garijo, Ver~Steeg, Szekely,
  Brekelmans, Kejriwal, Luo, and Huang]{gilp4ml}
Yolanda Gil, Ke-Thia Yao, Varun Ratnakar, Daniel Garijo, Greg Ver~Steeg, Pedro
  Szekely, Rob Brekelmans, Mayank Kejriwal, Fanghao Luo, and I-Hui Huang.
\newblock {P4ML}: A phased performance-based pipeline planner for automated
  machine learning.
\newblock In \emph{AutoML Workshop at ICML 2018}, 2018.

\bibitem[Giraud-Carrier(2008)]{giraud2008metalearning}
Christophe Giraud-Carrier.
\newblock Metalearning-a tutorial.
\newblock In \emph{Tutorial at the International Conference on Machine Learning
  and Applications}, pages 1--45, 2008.

\bibitem[Giraud-Carrier and Provost(2005)]{giraud2005toward}
Christophe Giraud-Carrier and Foster Provost.
\newblock Toward a justification of meta-learning: Is the no free lunch theorem
  a show-stopper.
\newblock In \emph{Proceedings of the ICML-2005 Workshop on Meta-learning},
  pages 12--19, 2005.

\bibitem[Golovin et~al.(2017)Golovin, Solnik, Moitra, Kochanski, Karro, and
  Sculley]{Golovin2017}
D.~Golovin, B.~Solnik, S.~Moitra, G.~Kochanski, J.~Karro, and D.~Sculley.
\newblock Google vizier: A service for black-box optimization.
\newblock In \emph{ICDM}, pages 1487--1495, 2017.

\bibitem[Gomes et~al.(2012)Gomes, Prud{\^e}ncio, Soares, Rossi, and
  Carvalho]{gomes2012combining}
Taciana~AF Gomes, Ricardo~BC Prud{\^e}ncio, Carlos Soares, Andr{\'e}~LD Rossi,
  and Andr{\'e} Carvalho.
\newblock Combining meta-learning and search techniques to select parameters
  for support vector machines.
\newblock \emph{Neurocomputing}, 75\penalty0 (1):\penalty0 3--13, 2012.

\bibitem[Grant et~al.(2018)Grant, Finn, Levine, Darrell, and
  Griffiths]{grant2018recasting}
Erin Grant, Chelsea Finn, Sergey Levine, Trevor Darrell, and Thomas Griffiths.
\newblock Recasting gradient-based meta-learning as hierarchical bayes.
\newblock \emph{arXiv preprint arXiv:1801.08930}, 2018.

\bibitem[Graves et~al.(2014)Graves, Wayne, and Danihelka]{graves2014neural}
Alex Graves, Greg Wayne, and Ivo Danihelka.
\newblock Neural turing machines.
\newblock \emph{arXiv preprint arXiv:1410.5401}, 2014.

\bibitem[Guerra et~al.(2008)Guerra, Prud{\^e}ncio, and
  Ludermir]{guerra2008predicting}
Silvio~B Guerra, Ricardo~BC Prud{\^e}ncio, and Teresa~B Ludermir.
\newblock Predicting the performance of learning algorithms using support
  vector machines as meta-regressors.
\newblock In \emph{ICANN}, pages 523--532, 2008.

\bibitem[Hengst(2002)]{Hengst02}
B.~Hengst.
\newblock {Discovering Hierarchy in Reinforcement Learning with HEXQ}.
\newblock In \emph{International Conference on Machine Learning}, pages
  243--250, 2002.

\bibitem[Hilario and Kalousis(2001)]{Hilario:2001p001}
M~Hilario and A~Kalousis.
\newblock Fusion of meta-knowledge and meta-data for case-based model
  selection.
\newblock \emph{Lecture Notes in Computer Science}, 2168:\penalty0 180--191,
  2001.

\bibitem[Ho and Basu(2002)]{Ho:2002}
Tin~Kam Ho and Mitra Basu.
\newblock Complexity measures of supervised classification problems.
\newblock \emph{Pattern Analysis and Machine Intellig.}, 24\penalty0
  (3):\penalty0 289--300, 2002.

\bibitem[Hochreiter et~al.(2001)Hochreiter, Younger, and
  Conwell]{hochreiter+01}
S.~Hochreiter, A.S. Younger, and P.R. Conwell.
\newblock Learning to learn using gradient descent.
\newblock In \emph{Lecture Notes on Computer Science, 2130}, pages 87--94,
  2001.

\bibitem[Hochreiter and Schmidhuber(1997)]{hochreiter1997long}
Sepp Hochreiter and J{\"u}rgen Schmidhuber.
\newblock Long short-term memory.
\newblock \emph{Neural computation}, 9\penalty0 (8):\penalty0 1735--1780, 1997.

\bibitem[Hutter et~al.(2014{\natexlab{a}})Hutter, Hoos, and
  Leyton-Brown]{fANOVA}
F.~Hutter, H.~Hoos, and K.~Leyton-Brown.
\newblock {An Efficient Approach for Assessing Hyperparameter Importance}.
\newblock In \emph{Proceedings of ICML}, 2014{\natexlab{a}}.

\bibitem[Hutter et~al.(2014{\natexlab{b}})Hutter, Xu, Hoos, and
  Leyton-Brown]{Hutter+2014}
F.~Hutter, L.~Xu, H.~Hoos, and K.~Leyton-Brown.
\newblock { Algorithm runtime prediction: Methods \& evaluation}.
\newblock \emph{Artificial Intelligence}, 206:\penalty0 79--111,
  2014{\natexlab{b}}.

\bibitem[Jones et~al.(1998)Jones, Schonlau, and Welch]{jones1998efficient}
Donald~R Jones, Matthias Schonlau, and William~J Welch.
\newblock Efficient global optimization of expensive black-box functions.
\newblock \emph{Journal of Global Optimization}, 13\penalty0 (4):\penalty0
  455--492, 1998.

\bibitem[Kalousis(2002)]{kalousis02}
A.~Kalousis.
\newblock \emph{Algorithm Selection via Meta-Learning}.
\newblock PhD thesis, University of Geneva, Department of Computer Science,
  2002.

\bibitem[Kalousis and Hilario(2003)]{Kalousis:2003p335}
A~Kalousis and M~Hilario.
\newblock Representational issues in meta-learning.
\newblock \emph{Proceedings of ICML 2003}, pages 313--320, 2003.

\bibitem[Kalousis and Hilario(2001)]{Kalousis2001a}
Alexandros Kalousis and Melanie Hilario.
\newblock Model selection via meta-learning: a comparative study.
\newblock \emph{Intl Journ. on Artificial Intelligence Tools}, 10\penalty0
  (4):\penalty0 525--554, 2001.

\bibitem[Kendall(1938)]{kendall1938new}
Maurice~G Kendall.
\newblock A new measure of rank correlation.
\newblock \emph{Biometrika}, 30\penalty0 (1/2):\penalty0 81--93, 1938.

\bibitem[Kietz et~al.(2012)Kietz, Serban, Bernstein, and
  Fischer]{kietz2012designing}
J{\"o}rg-Uwe Kietz, Floarea Serban, Abraham Bernstein, and Simon Fischer.
\newblock Designing {KDD}-workflows via {HTN}-planning for intelligent
  discovery assistance.
\newblock In \emph{5th Planning to Learn Workshop at ECAI 2012}, 2012.

\bibitem[Kim et~al.(2017)Kim, Kim, and Choi]{kim2017}
J.~Kim, S.~Kim, and S.~Choi.
\newblock Learning to warm-start {B}ayesian hyperparameter optimization.
\newblock \emph{arXiv preprint arXiv:1710.06219}, 2017.

\bibitem[Kohavi and John(1995)]{kohavi1995automatic}
Ron Kohavi and George~H John.
\newblock Automatic parameter selection by minimizing estimated error.
\newblock In \emph{Proceedings of the International Conference Machine
  Learning}, pages 304--312, 1995.

\bibitem[K{\"o}pf and Iglezakis(2002)]{Kopf:2002p5864}
C~K{\"o}pf and I~Iglezakis.
\newblock Combination of task description strategies and case base properties
  for meta-learning.
\newblock \emph{ECML/PKDD Workshop on Integration and Collaboration Aspects of
  Data Mining}, pages 65--76, 2002.

\bibitem[K\"{o}pf et~al.(2000)K\"{o}pf, Taylor, and Keller]{kopf+00}
C.~K\"{o}pf, C.~Taylor, and J.~Keller.
\newblock Meta-analysis: From data characterization for meta-learning to
  meta-regression.
\newblock In \emph{{PKDD} Workshop on Data Mining, Decision Support,
  Meta-Learning and {ILP}.}, pages 15--26, 2000.

\bibitem[Krizhevsky et~al.(2012)Krizhevsky, Sutskever, and
  Hinton]{krizhevsky2012imagenet}
Alex Krizhevsky, Ilya Sutskever, and Geoffrey~E Hinton.
\newblock Imagenet classification with deep convolutional neural networks.
\newblock In \emph{Advances in neural information processing systems}, pages
  1097--1105, 2012.

\bibitem[Kuba et~al.(2002)Kuba, Brazdil, Soares, and Woznica]{kuba+02a}
P.~Kuba, P.~Brazdil, C.~Soares, and A.~Woznica.
\newblock Exploiting sampling and meta-learning for parameter setting support
  vector machines.
\newblock In \emph{Proceedings of IBERAMIA 2002}, pages 217--225, 2002.

\bibitem[Kullback and Leibler(1951)]{kullback1951information}
Solomon Kullback and Richard~A Leibler.
\newblock On information and sufficiency.
\newblock \emph{The annals of mathematical statistics}, 22\penalty0
  (1):\penalty0 79--86, 1951.

\bibitem[Lacoste et~al.(2014)Lacoste, Marchand, Laviolette, and
  Larochelle]{lacoste2014agnostic}
Alexandre Lacoste, Mario Marchand, Fran{\c{c}}ois Laviolette, and Hugo
  Larochelle.
\newblock Agnostic {B}ayesian learning of ensembles.
\newblock In \emph{ICML}, pages 611--619, 2014.

\bibitem[Lake et~al.(2017)Lake, Ullman, Tenenbaum, and
  Gershman]{lake2017building}
Brenden~M Lake, Tomer~D Ullman, Joshua~B Tenenbaum, and Samuel~J Gershman.
\newblock Building machines that learn and think like people.
\newblock \emph{Beh. and Brain Sc.}, 40, 2017.

\bibitem[Leite and Brazdil(2005)]{Leite:2005p725}
R~Leite and P~Brazdil.
\newblock Predicting relative performance of classifiers from samples.
\newblock \emph{Proceedings of ICML}, pages 497--504, 2005.

\bibitem[Leite and Brazdil(2007)]{Leite:2007p6022}
R~Leite and P~Brazdil.
\newblock An iterative process for building learning curves and predicting
  relative performance of classifiers.
\newblock \emph{Lecture Notes in Computer Science}, 4874:\penalty0 87--98,
  2007.

\bibitem[Leite et~al.(2012)Leite, Brazdil, and Vanschoren]{leite+12}
R.~Leite, P.~Brazdil, and J.~Vanschoren.
\newblock {Selecting Classification Algorithms with Active Testing}.
\newblock \emph{Lecture Notes in Artif. Intel.}, 10934:\penalty0 117--131,
  2012.

\bibitem[Leite and Brazdil(2010)]{Leite:2010}
Rui Leite and Pavel Brazdil.
\newblock Active testing strategy to predict the best classification algorithm
  via sampling and metalearning.
\newblock In \emph{ECAI 2010}, pages 309--314, 2010.

\bibitem[Lemke et~al.(2015)Lemke, Budka, and Gabrys]{lemke+15}
C.~Lemke, M.~Budka, and B.~Gabrys.
\newblock Metalearning: a survey of trends and technologies.
\newblock \emph{Artificial intelligence review}, 44\penalty0 (1):\penalty0
  117--130, 2015.

\bibitem[Ler et~al.(2005)Ler, Koprinska, and Chawla]{Ler:2005p1680}
Daren Ler, Irena Koprinska, and Sanjay Chawla.
\newblock Utilizing regression-based landmarkers within a meta-learning
  framework for algorithm selection.
\newblock \emph{Technical Report 569. University of Sydney}, pages 44--51,
  2005.

\bibitem[Li and Malik(2016)]{li2016learning}
Ke~Li and Jitendra Malik.
\newblock Learning to optimize.
\newblock \emph{arXiv preprint arXiv:1606.01885}, 2016.

\bibitem[Li and Malik(2017)]{li2017learning}
Ke~Li and Jitendra Malik.
\newblock Learning to optimize neural nets.
\newblock \emph{arXiv preprint arXiv:1703.00441}, 2017.

\bibitem[Lin(2010)]{Lin2010}
S.~Lin.
\newblock {Rank aggregation methods}.
\newblock \emph{WIREs Computational Statistics}, 2:\penalty0 555--570, 2010.

\bibitem[Lindner and Studer(1999)]{lindner+99}
G.~Lindner and R.~Studer.
\newblock {AST}: Support for algorithm selection with a {CBR} approach.
\newblock In \emph{ICML Workshop on Recent Advances in Meta-Learning and Future
  Work}, pages 38--47. J. Stefan Institute, 1999.

\bibitem[Lorena et~al.(2018)Lorena, Maciel, de~Miranda, Costa, and
  Prud{\^{e}}ncio]{Lorena:2018}
Ana~Carolina Lorena, Aron~I. Maciel, P{\'{e}}ricles B.~C. de~Miranda, Ivan~G.
  Costa, and Ricardo B.~C. Prud{\^{e}}ncio.
\newblock Data complexity meta-features for regression problems.
\newblock \emph{Machine Learning}, 107\penalty0 (1):\penalty0 209--246, 2018.
\newblock \doi{10.1007/s10994-017-5681-1}.

\bibitem[Luo(2016)]{luo2016review}
Gang Luo.
\newblock A review of automatic selection methods for machine learning
  algorithms and hyper-parameter values.
\newblock \emph{Network Modeling Analysis in Health Informatics and
  Bioinformatics}, 5\penalty0 (1):\penalty0 18, 2016.

\bibitem[Mantovani et~al.(2015{\natexlab{a}})Mantovani, Rossi, Vanschoren,
  Bischl, and Carvalho]{mantovani2015tune}
Rafael~G Mantovani, Andr{\'e}~LD Rossi, Joaquin Vanschoren, Bernd Bischl, and
  Andr{\'e}~CPLF Carvalho.
\newblock To tune or not to tune: recommending when to adjust {SVM}
  hyper-parameters via meta-learning.
\newblock In \emph{Proceedings of IJCNN}, pages 1--8, 2015{\natexlab{a}}.

\bibitem[Mantovani et~al.(2016)Mantovani, Horv{\'a}th, Cerri, Vanschoren, and
  de~Carvalho]{mantovani2016hyper}
Rafael~G Mantovani, Tom{\'a}{\v{s}} Horv{\'a}th, Ricardo Cerri, Joaquin
  Vanschoren, and Andr{\'e}~CPLF de~Carvalho.
\newblock Hyper-parameter tuning of a decision tree induction algorithm.
\newblock In \emph{Brazilian Conference on Intelligent Systems}, pages 37--42,
  2016.

\bibitem[Mantovani et~al.(2015{\natexlab{b}})Mantovani, Rossi, Vanschoren, and
  Carvalho]{mantovani2015meta}
Rafael~Gomes Mantovani, Andr{\'e}~LD Rossi, Joaquin Vanschoren, and
  Andr{\'e}~Carlos Carvalho.
\newblock Meta-learning recommendation of default hyper-parameter values for
  {SVM}s in classifications tasks.
\newblock In \emph{ECML PKDD Workshop on Meta-Learning and Algorithm
  Selection}, 2015{\natexlab{b}}.

\bibitem[Mantovani(2018)]{Mantovani:2018}
R.G. Mantovani.
\newblock \emph{Use of meta-learning for hyperparameter tuning of
  classification problems}.
\newblock PhD thesis, University of Sao Carlos, Brazil, 2018.

\bibitem[Michie et~al.(1994)Michie, Spiegelhalter, Taylor, and
  Campbell]{Michie1994}
Donald Michie, David~J. Spiegelhalter, Charles~C. Taylor, and John Campbell.
\newblock \emph{{Machine Learning, Neural and Statistical Classification}}.
\newblock Ellis Horwood, 1994.

\bibitem[Miranda and Prud{\^e}ncio(2013)]{Miranda:2013}
P.B.C. Miranda and R.B.C. Prud{\^e}ncio.
\newblock Active testing for {SVM} parameter selection.
\newblock In \emph{Proceedings of IJCNN}, pages 1--8, 2013.

\bibitem[Mishra et~al.(2018)Mishra, Rohaninejad, Chen, and
  Abbeel]{mishra2018simple}
Nikhil Mishra, Mostafa Rohaninejad, Xi~Chen, and Pieter Abbeel.
\newblock A simple neural attentive meta-learner.
\newblock In \emph{Proceedings of ICLR}, 2018.

\bibitem[Misir and Sebag(2013)]{misir:hal-00922840}
Mustafa Misir and Mich{\`e}le Sebag.
\newblock {Algorithm Selection as a Collaborative Filtering Problem}.
\newblock Research report, INRIA, 2013.

\bibitem[M{\i}s{\i}r and Sebag(2017)]{misir2017alors}
Mustafa M{\i}s{\i}r and Mich{\`e}le Sebag.
\newblock Alors: An algorithm recommender system.
\newblock \emph{Artificial Intelligence}, 244:\penalty0 291--314, 2017.

\bibitem[Nadaraya(1964)]{nadaraya1964estimating}
Elizbar~A Nadaraya.
\newblock On estimating regression.
\newblock \emph{Theory of Probability \& Its Applications}, 9\penalty0
  (1):\penalty0 141--142, 1964.

\bibitem[Nguyen et~al.(2014)Nguyen, Hilario, and Kalousis]{nguyen2014using}
Phong Nguyen, Melanie Hilario, and Alexandros Kalousis.
\newblock Using meta-mining to support data mining workflow planning and
  optimization.
\newblock \emph{Journal of Artificial Intelligence Research}, 51:\penalty0
  605--644, 2014.

\bibitem[Nichol et~al.(2018)Nichol, Achiam, and Schulman]{Nichol2018}
A.~Nichol, J.~Achiam, and J.~Schulman.
\newblock On first-order meta-learning algorithms.
\newblock \emph{arXiv}, 1803.02999v2, 2018.

\bibitem[Niculescu-Mizil and Caruana(2005)]{Niculescu05}
A.~Niculescu-Mizil and R.~Caruana.
\newblock {Learning the Structure of Related Tasks}.
\newblock In \emph{Proceedings of NIPS Workshop on Inductive Transfer}, 2005.

\bibitem[Nisioti et~al.(2018)Nisioti, Chatzidimitriou, and
  Symeonidis]{Nisioti2018}
E.~Nisioti, K.~Chatzidimitriou, and A~Symeonidis.
\newblock Predicting hyperparameters from meta-features in binary
  classification problems.
\newblock In \emph{AutoML Workshop at ICML}, 2018.

\bibitem[Olier et~al.(2018)Olier, Sadawi, Bickerton, Vanschoren, Grosan,
  Soldatova, and King]{Olier2018}
I.~Olier, N.~Sadawi, G.R. Bickerton, J.~Vanschoren, C.~Grosan, L.~Soldatova,
  and R.D. King.
\newblock {Meta-QSAR: learning how to learn QSARs}.
\newblock \emph{Machine Learning}, 107:\penalty0 285--311, 2018.

\bibitem[Olson et~al.(2016)Olson, Bartley, Urbanowicz, and
  Moore]{olson2016evaluation}
Randal~S Olson, Nathan Bartley, Ryan~J Urbanowicz, and Jason~H Moore.
\newblock Evaluation of a tree-based pipeline optimization tool for automating
  data science.
\newblock In \emph{Proceedings of GECCO}, pages 485--492, 2016.

\bibitem[Pan and Yang(2010)]{pan2010survey}
Sinno~Jialin Pan and Qiang Yang.
\newblock A survey on transfer learning.
\newblock \emph{IEEE Transactions on knowledge and data engineering},
  22\penalty0 (10):\penalty0 1345--1359, 2010.

\bibitem[Pang et~al.(2018)Pang, Dong, Wu, and Hospedales]{Pang2018}
K~Pang, M.~Dong, Y.~Wu, and T.~Hospedales.
\newblock Meta-learning transferable active learning policies by deep
  reinforcement learning.
\newblock In \emph{AutoML Workshop at ICML}, 2018.

\bibitem[Peng et~al.(2002)Peng, Flach, Soares, and Brazdil]{Peng:2002p705}
Y~Peng, P~Flach, C~Soares, and P~Brazdil.
\newblock Improved dataset characterisation for meta-learning.
\newblock \emph{Lecture Notes in Com. Sc.}, 2534:\penalty0 141--152, 2002.

\bibitem[Perrone et~al.(2017)Perrone, Jenatton, Seeger, and
  Archambeau]{perrone2017multiple}
Valerio Perrone, Rodolphe Jenatton, Matthias Seeger, and Cedric Archambeau.
\newblock Multiple adaptive {B}ayesian linear regression for scalable
  {B}ayesian optimization with warm start.
\newblock \emph{arXiv preprint arXiv:1712.02902}, 2017.

\bibitem[Pfahringer et~al.(2000)Pfahringer, Bensusan, and
  Giraud-Carrier]{Pfahringer:2000p553}
Bernhard Pfahringer, Hilan Bensusan, and Christophe~G. Giraud-Carrier.
\newblock Meta-learning by landmarking various learning algorithms.
\newblock In \emph{17th International Conference on Machine Learning (ICML)},
  pages 743 -- 750, 2000.

\bibitem[Pinto et~al.(2016)Pinto, Soares, and Mendes-Moreira]{Pinto2016}
F{\'a}bio Pinto, Carlos Soares, and Jo{\~a}o Mendes-Moreira.
\newblock Towards automatic generation of metafeatures.
\newblock In \emph{Proceedings of PAKDD}, pages 215--226, 2016.

\bibitem[Pinto et~al.(2017)Pinto, Cerqueira, Soares, and
  Mendes{-}Moreira]{Pinto:2017}
F{\'{a}}bio Pinto, V{\'{\i}}tor Cerqueira, Carlos Soares, and Jo{\~{a}}o
  Mendes{-}Moreira.
\newblock {autoBagging}: Learning to rank bagging workflows with metalearning.
\newblock \emph{arXiv}, 1706.09367, 2017.

\bibitem[Post et~al.(2016)Post, van~der Putten, and van Rijn]{Post2016}
Martijn~J. Post, Peter van~der Putten, and Jan~N. van Rijn.
\newblock {Does Feature Selection Improve Classification? A Large Scale
  Experiment in OpenML}.
\newblock In \emph{Advances in Intelligent Data Analysis XV}, pages 158--170,
  2016.

\bibitem[Priya et~al.(2012)Priya, {De Souza}, Rossi, and Carvalho]{Priya:2012}
Rattan Priya, Bruno~F. {De Souza}, Andre Rossi, and Andre Carvalho.
\newblock {Using genetic algorithms to improve prediction of execution times of
  ML tasks}.
\newblock In \emph{Lecture Notes in Comp. Science}, volume 7208, pages
  196--207, 2012.

\bibitem[Probst et~al.(2018)Probst, Bischl, and Boulesteix]{Probst2018}
P.~Probst, B.~Bischl, and A.-L. Boulesteix.
\newblock {Tunability: Importance of hyperparameters of machine learning
  algorithms}.
\newblock \emph{ArXiv 1802.09596}, 2018.

\bibitem[Provost et~al.(1999)Provost, Jensen, and Oates]{provost1999efficient}
Foster Provost, David Jensen, and Tim Oates.
\newblock Efficient progressive sampling.
\newblock In \emph{Proceedings of the fifth ACM SIGKDD international conference
  on Knowledge discovery and data mining}, pages 23--32, 1999.

\bibitem[Prud{\^e}ncio and Ludermir(2004)]{Prudencio:2004p6308}
R~Prud{\^e}ncio and T~Ludermir.
\newblock Meta-learning approaches to selecting time series models.
\newblock \emph{Neurocomputing}, 61:\penalty0 121--137, 2004.

\bibitem[Raina et~al.(2006)Raina, Ng, and Koller]{Raina05}
R.~Raina, A.~Y. Ng, and D.~Koller.
\newblock {Transfer Learning by Constructing Informative Priors}.
\newblock In \emph{Proceedings of ICML}, 2006.

\bibitem[Ramachandran et~al.(2018{\natexlab{a}})Ramachandran, Gupta, Rana, and
  Venkatesh]{Ramachandran2018}
Anil Ramachandran, Sunil Gupta, Santu Rana, and Svetha Venkatesh.
\newblock Selecting optimal source for transfer learning in {B}ayesian
  optimisation.
\newblock In \emph{Proceedings of PRICAI}, pages 42--56, 2018{\natexlab{a}}.

\bibitem[Ramachandran et~al.(2018{\natexlab{b}})Ramachandran, Gupta, Rana, and
  Venkatesh]{Ramachandran2018b}
Anil Ramachandran, Sunil Gupta, Santu Rana, and Svetha Venkatesh.
\newblock Information-theoretic transfer learning framework for {B}ayesian
  optimisation.
\newblock In \emph{Proceedings of ECMLPKDD}, 2018{\natexlab{b}}.

\bibitem[Rasmussen(2004)]{rasmussen2004gaussian}
Carl~Edward Rasmussen.
\newblock Gaussian processes in machine learning.
\newblock In \emph{Advanced lectures on machine learning}, pages 63--71.
  Springer, 2004.

\bibitem[Ravi and Larochelle(2017)]{ravi2016optimization}
Sachin Ravi and Hugo Larochelle.
\newblock Optimization as a model for few-shot learning.
\newblock In \emph{Proceedings of ICLR}, 2017.

\bibitem[Reed et~al.(2017)Reed, Chen, Paine, Oord, Eslami, Rezende, Vinyals,
  and de~Freitas]{reed2017few}
Scott Reed, Yutian Chen, Thomas Paine, A{\"a}ron van~den Oord, SM~Eslami,
  Danilo Rezende, Oriol Vinyals, and Nando de~Freitas.
\newblock Few-shot autoregressive density estimation: Towards learning to learn
  distributions.
\newblock \emph{arXiv preprint arXiv:1710.10304}, 2017.

\bibitem[Reif et~al.(2011)Reif, Shafait, and Dengel]{Reif2011}
Matthias Reif, Faisal Shafait, and Andreas Dengel.
\newblock Prediction of classifier training time including parameter
  optimization.
\newblock In \emph{Proc. of GfKI}, pages 260 -- 271, 2011.

\bibitem[Reif et~al.(2012)Reif, Shafait, and Dengel]{reif2012meta}
Matthias Reif, Faisal Shafait, and Andreas Dengel.
\newblock Meta-learning for evolutionary parameter optimization of classifiers.
\newblock \emph{Machine learning}, 87\penalty0 (3):\penalty0 357--380, 2012.

\bibitem[Reif et~al.(2014)Reif, Shafait, Goldstein, Breuel, and
  Dengel]{Reif2014}
Matthias Reif, Faisal Shafait, Markus Goldstein, Thomas Breuel, and Andreas
  Dengel.
\newblock Automatic classifier selection for non-experts.
\newblock \emph{Pattern Analysis and Applications}, 17\penalty0 (1):\penalty0
  83 -- 96, 2014.

\bibitem[Ren et~al.(2018)Ren, Triantafillou, Ravi, Snell, Swersky, Tenenbaum,
  Larochelle, and Zemel]{ren2018meta}
Mengye Ren, Eleni Triantafillou, Sachin Ravi, Jake Snell, Kevin Swersky,
  Joshua~B Tenenbaum, Hugo Larochelle, and Richard~S Zemel.
\newblock Meta-learning for semi-supervised few-shot classification.
\newblock \emph{arXiv 1803.00676}, 2018.

\bibitem[Rendle(2010)]{Rendle2010}
S~Rendle.
\newblock Factorization machines.
\newblock In \emph{ICDM 2015}, pages 995--1000, 2010.

\bibitem[Ridd and Giraud-Carrier(2014)]{Ridd:2014}
Parker Ridd and Christophe Giraud-Carrier.
\newblock Using metalearning to predict when parameter optimization is likely
  to improve classification accuracy.
\newblock In \emph{ECAI Workshop on Meta-learning and Algorithm Selection},
  pages 18--23, 2014.

\bibitem[Rivolli et~al.(2018)Rivolli, Garcia, Soares, Vanschoren, and
  de~Carvalho]{rivolli2018}
A.~Rivolli, L.P.F. Garcia, C.~Soares, J.~Vanschoren, and A.C.P.L.F.
  de~Carvalho.
\newblock Towards reproducible empirical research in meta-learning.
\newblock \emph{arXiv preprint}, 1808.10406, 2018.

\bibitem[Robbins(1985)]{robbins1985some}
Herbert Robbins.
\newblock Some aspects of the sequential design of experiments.
\newblock In \emph{Herbert Robbins Selected Papers}, pages 169--177. Springer,
  1985.

\bibitem[Rosenstein et~al.(2005)Rosenstein, Marx, and Kaelbling]{Rosenstein05}
M.~T. Rosenstein, Z.~Marx, and L.~P. Kaelbling.
\newblock {To Transfer or Not To Transfer}.
\newblock In \emph{NIPS Workshop on transfer learning}, 2005.

\bibitem[Rousseeuw and Hubert(2011)]{Rousseeuw2011}
Peter~J. Rousseeuw and Mia Hubert.
\newblock Robust statistics for outlier detection.
\newblock \emph{Wiley Interdisciplinary Reviews: Data Mining and Knowledge
  Discovery}, 1\penalty0 (1):\penalty0 73 -- 79, 2011.

\bibitem[Runarsson and Jonsson(2000)]{runarsson2000evolution}
Thomas~Philip Runarsson and Magnus~Thor Jonsson.
\newblock Evolution and design of distributed learning rules.
\newblock In \emph{IEEE Symposium on Combinations of Evolutionary Computation
  and Neural Networks}, pages 59--63, 2000.

\bibitem[Salama et~al.(2013)Salama, Hassanien, and Revett]{Salama2013}
Mostafa~A. Salama, Aboul~Ella Hassanien, and Kenneth Revett.
\newblock Employment of neural network and rough set in meta-learning.
\newblock \emph{Memetic Comp.}, 5\penalty0 (3):\penalty0 165--177, 2013.

\bibitem[Sanders and Giraud-Carrier(2017)]{Sanders8215600}
S.~Sanders and C.~Giraud-Carrier.
\newblock Informing the use of hyperparameter optimization through
  metalearning.
\newblock In \emph{Proc. ICDM}, pages 1051--1056, 2017.

\bibitem[Santoro et~al.(2016{\natexlab{a}})Santoro, Bartunov, Botvinick,
  Wierstra, and Lillicrap]{santoro2016meta}
Adam Santoro, Sergey Bartunov, Matthew Botvinick, Daan Wierstra, and Timothy
  Lillicrap.
\newblock Meta-learning with memory-augmented neural networks.
\newblock In \emph{International conference on machine learning}, pages
  1842--1850, 2016{\natexlab{a}}.

\bibitem[Santoro et~al.(2016{\natexlab{b}})Santoro, Bartunov, Botvinick,
  Wierstra, and Lillicrap]{santoro2016one}
Adam Santoro, Sergey Bartunov, Matthew Botvinick, Daan Wierstra, and Timothy
  Lillicrap.
\newblock One-shot learning with memory-augmented neural networks.
\newblock \emph{arXiv preprint arXiv:1605.06065}, 2016{\natexlab{b}}.

\bibitem[Schilling et~al.(2015)Schilling, Wistuba, Drumond, and
  Schmidt-Thieme]{Schilling2015}
N.~Schilling, M.~Wistuba, L.~Drumond, and L.~Schmidt-Thieme.
\newblock Hyperparameter optimization with factorized multilayer perceptrons.
\newblock In \emph{Proceedings of ECML PKDD}, pages 87--103, 2015.

\bibitem[Schmidhuber(1992)]{schmidhuber1992learning}
J{\"u}rgen Schmidhuber.
\newblock Learning to control fast-weight memories: An alternative to dynamic
  recurrent networks.
\newblock \emph{Neural Comp.}, 4\penalty0 (1):\penalty0 131--139, 1992.

\bibitem[Schmidhuber(1993)]{schmidhuber1993neural}
J{\"u}rgen Schmidhuber.
\newblock A neural network that embeds its own meta-levels.
\newblock In \emph{Proceedings of ICNN}, pages 407--412, 1993.

\bibitem[Schmidhuber et~al.(1997)Schmidhuber, Zhao, and
  Wiering]{schmidhuber1997shifting}
J{\"u}rgen Schmidhuber, Jieyu Zhao, and Marco Wiering.
\newblock Shifting inductive bias with success-story algorithm, adaptive levin
  search, and incremental self-improvement.
\newblock \emph{Machine Learning}, 28\penalty0 (1):\penalty0 105--130, 1997.

\bibitem[Schoenfeld et~al.(2018)Schoenfeld, Giraud-Carrier, Poggeman,
  Christensen, and Seppi]{Schoenfeld2018}
B.~Schoenfeld, C.~Giraud-Carrier, M.~Poggeman, J.~Christensen, and K.~Seppi.
\newblock Feature selection for high-dimensional data: A fast correlation-based
  filter solution.
\newblock In \emph{AutoML Workshop at ICML}, 2018.

\bibitem[Serban et~al.(2013)Serban, Vanschoren, Kietz, and
  Bernstein]{Serban2013}
F.~Serban, J.~Vanschoren, J.U. Kietz, and A.A Bernstein.
\newblock A survey of intelligent assistants for data analysis.
\newblock \emph{ACM Computing Surveys}, 45\penalty0 (3):\penalty0 Art.31, 2013.

\bibitem[Sharif~Razavian et~al.(2014)Sharif~Razavian, Azizpour, Sullivan, and
  Carlsson]{sharif2014cnn}
Ali Sharif~Razavian, Hossein Azizpour, Josephine Sullivan, and Stefan Carlsson.
\newblock Cnn features off-the-shelf: an astounding baseline for recognition.
\newblock In \emph{Proceedings of CVPR 2014}, pages 806--813, 2014.

\bibitem[Sharkey and Sharkey(1993)]{Sharkey93}
N.~E. Sharkey and A.~J.~C. Sharkey.
\newblock {Adaptive Generalization}.
\newblock \emph{Artificial Intelligence Review}, 7:\penalty0 313--328, 1993.

\bibitem[Smith-Miles(2009)]{Smith-Miles2008}
Kate~A. Smith-Miles.
\newblock Cross-disciplinary perspectives on meta-learning for algorithm
  selection.
\newblock \emph{ACM Computing Surveys}, 41\penalty0 (1):\penalty0 1 -- 25,
  2009.

\bibitem[Snell et~al.(2017)Snell, Swersky, and Zemel]{snell2017prototypical}
Jake Snell, Kevin Swersky, and Richard Zemel.
\newblock Prototypical networks for few-shot learning.
\newblock In \emph{Neural Information Processing Systems}, pages 4077--4087,
  2017.

\bibitem[Soares et~al.(2001)Soares, Petrak, and Brazdil]{Soares:2001p708}
C~Soares, J~Petrak, and P~Brazdil.
\newblock Sampling based relative landmarks: Systematically testdriving
  algorithms before choosing.
\newblock \emph{Lecture Notes in Computer Science}, 3201:\penalty0 250--261,
  2001.

\bibitem[Soares et~al.(2004)Soares, Brazdil, and Kuba]{soares+04}
C.~Soares, P.~Brazdil, and P.~Kuba.
\newblock A meta-learning method to select the kernel width in support vector
  regression.
\newblock \emph{Mach. Learn.}, 54:\penalty0 195--209, 2004.

\bibitem[Soares et~al.(2009)Soares, Ludermir, and Carvalho]{Soares:2009p9657}
C~Soares, T~Ludermir, and F~De Carvalho.
\newblock An analysis of meta-learning techniques for ranking clustering
  algorithms applied to artificial data.
\newblock \emph{Lecture Notes in Computer Science}, 5768:\penalty0 131--140,
  2009.

\bibitem[Springenberg et~al.(2016)Springenberg, Klein, Falkner, and
  Hutter]{Springenberg2016}
J.~Springenberg, A.~Klein, S.~Falkner, and Frank Hutter.
\newblock {B}ayesian optimization with robust {B}ayesian neural networks.
\newblock In \emph{Advances in Neural Information Processing Systems}, 2016.

\bibitem[Stern et~al.(2010)Stern, Samulowitz, Herbrich, Graepel, Pulina, and
  Tacchella]{stern2010collaborative}
David~H Stern, Horst Samulowitz, Ralf Herbrich, Thore Graepel, Luca Pulina, and
  Armando Tacchella.
\newblock Collaborative expert portfolio management.
\newblock In \emph{Proceedings of AAAI}, pages 179--184, 2010.

\bibitem[Strang et~al.(2018)Strang, van~der Putten, van Rijn, and
  Hutter]{Strang2018}
Benjamin Strang, Peter van~der Putten, Jan~N. van Rijn, and Frank Hutter.
\newblock {Don't Rule Out Simple Models Prematurely}.
\newblock In \emph{Adv. in Intelligent Data Analysis}, 2018.

\bibitem[Sun et~al.(2013)Sun, Pfahringer, and Mayo]{Sun2013a}
Q.~Sun, B.~Pfahringer, and M.~Mayo.
\newblock {Towards a Framework for Designing Full Model Selection and
  Optimization Systems}.
\newblock In \emph{International Workshop on Multiple Classifier Systems},
  pages 259--270, 2013.

\bibitem[Sun and Pfahringer(2013)]{Sun:2013}
Quan Sun and Bernhard Pfahringer.
\newblock Pairwise meta-rules for better meta-learning-based algorithm ranking.
\newblock \emph{Machine Learning}, 93\penalty0 (1):\penalty0 141--161, 2013.

\bibitem[Swersky et~al.(2013)Swersky, Snoek, and Adams]{swersky2013multi}
Kevin Swersky, Jasper Snoek, and Ryan~P Adams.
\newblock Multi-task {B}ayesian optimization.
\newblock In \emph{Adv. in neural information processing systems}, pages
  2004--2012, 2013.

\bibitem[Swersky et~al.(2014)Swersky, Snoek, and Adams]{swersky2014freeze}
Kevin Swersky, Jasper Snoek, and Ryan~Prescott Adams.
\newblock Freeze-thaw bayesian optimization.
\newblock \emph{arXiv preprint arXiv:1406.3896}, 2014.

\bibitem[Thompson(1933)]{thompson1933likelihood}
William~R Thompson.
\newblock On the likelihood that one unknown probability exceeds another in
  view of the evidence of two samples.
\newblock \emph{Biometrika}, 25\penalty0 (3/4):\penalty0 285--294, 1933.

\bibitem[Thrun(1998)]{Thrun98}
S.~Thrun.
\newblock {Lifelong Learning Algorithms}.
\newblock In \emph{Learning to Learn}, chapter~8, pages 181--209. Kluwer
  Academic Publishers, MA, 1998.

\bibitem[Thrun and Mitchell(1995)]{Thrun95A}
S.~Thrun and T.~Mitchell.
\newblock {Learning One More Thing}.
\newblock In \emph{Proceedings of IJCAI}, pages 1217--1223, 1995.

\bibitem[Thrun and Pratt(1998)]{Thrun98A}
S.~Thrun and L.~Pratt.
\newblock {Learning to Learn: Introduction and Overview}.
\newblock In \emph{Learning to Learn}, pages 3--17. Kluwer, 1998.

\bibitem[Todorovski and Dzeroski(1999)]{Todorovski:1999p5595}
L~Todorovski and S~Dzeroski.
\newblock Experiments in meta-level learning with {ILP}.
\newblock \emph{Lecture Notes in Computer Science}, 1704:\penalty0 98--106,
  1999.

\bibitem[Todorovski et~al.(2000)Todorovski, Brazdil, and
  Soares]{Todorovski:2000p5454}
L~Todorovski, P~Brazdil, and C~Soares.
\newblock Report on the experiments with feature selection in meta-level
  learning.
\newblock \emph{PKDD 2000 Workshop on Data mining, Decision support,
  Meta-learning and ILP}, pages 27--39, 2000.

\bibitem[Todorovski et~al.(2002)Todorovski, Blockeel, and
  {D\v{z}eroski}]{todorovski+02}
L.~Todorovski, H.~Blockeel, and S.~{D\v{z}eroski}.
\newblock Ranking with predictive clustering trees.
\newblock \emph{Lecture Notes in Artificial Intelligence}, 2430:\penalty0
  444--455, 2002.

\bibitem[van Rijn et~al.(2015)van Rijn, Abdulrahman, Brazdil, and
  Vanschoren]{vanrijn2015}
J.~van Rijn, S.~Abdulrahman, P.~Brazdil, and J.~Vanschoren.
\newblock {Fast Algorithm Selection Using Learning Curves}.
\newblock In \emph{Proceedings of IDA}, 2015.

\bibitem[van Rijn et~al.(2018)van Rijn, Holmes, Pfahringer, and
  Vanschoren]{vanRijn2018}
J.~van Rijn, G.~Holmes, B.~Pfahringer, and J.~Vanschoren.
\newblock {The Online Performance Estimation Framework. Heterogeneous Ensemble
  Learning for Data Streams.}
\newblock \emph{Machine Learning}, 107:\penalty0 149--176, 2018.

\bibitem[van Rijn and Hutter(2018)]{vanRijn2018b}
J.~N. van Rijn and Frank Hutter.
\newblock {Hyperparameter importance across datasets}.
\newblock In \emph{Proceedings of KDD}, pages 2367--2376, 2018.

\bibitem[van Rijn et~al.(2014)van Rijn, Holmes, Pfahringer, and
  Vanschoren]{van2014algorithm}
Jan~N van Rijn, Geoffrey Holmes, Bernhard Pfahringer, and Joaquin Vanschoren.
\newblock Algorithm selection on data streams.
\newblock In \emph{Discovery Science}, pages 325--336, 2014.

\bibitem[Vanschoren et~al.(2014)Vanschoren, van Rijn, Bischl, and
  Torgo]{Vanschoren2014}
J.~Vanschoren, J.~N. van Rijn, B.~Bischl, and L.~Torgo.
\newblock {OpenML: networked science in machine learning}.
\newblock \emph{ACM SIGKDD Explorations Newsletter}, 15\penalty0 (2):\penalty0
  49--60, 2014.

\bibitem[Vanschoren(2010)]{Vanschoren2010}
Joaquin Vanschoren.
\newblock \emph{Understanding Machine Learning Performance with Experiment
  Databases}.
\newblock PhD thesis, Leuven Univeristy, 2010.

\bibitem[Vanschoren et~al.(2012)Vanschoren, Blockeel, Pfahringer, and
  Holmes]{vanschoren2012experiment}
Joaquin Vanschoren, Hendrik Blockeel, Bernhard Pfahringer, and Geoffrey Holmes.
\newblock Experiment databases.
\newblock \emph{Machine Learning}, 87\penalty0 (2):\penalty0 127--158, 2012.

\bibitem[Vartak et~al.(2017)Vartak, Thiagarajan, Miranda, Bratman, and
  Larochelle]{vartak2017meta}
Manasi Vartak, Arvind Thiagarajan, Conrado Miranda, Jeshua Bratman, and Hugo
  Larochelle.
\newblock A meta-learning perspective on cold-start recommendations for items.
\newblock In \emph{Advances in Neural Information Processing Systems}, pages
  6904--6914, 2017.

\bibitem[Vilalta(1999)]{Vilalta:1999p5745}
R~Vilalta.
\newblock Understanding accuracy performance through concept characterization
  and algorithm analysis.
\newblock \emph{ICML Workshop on Recent Advances in Meta-Learning and Future
  Work}, 1999.

\bibitem[Vilalta and Drissi(2002)]{Vilalta:2002p5805}
R~Vilalta and Y~Drissi.
\newblock A characterization of difficult problems in classification.
\newblock \emph{Proceedings of ICMLA}, 2002.

\bibitem[Vinyals et~al.(2016)Vinyals, Blundell, Lillicrap, Wierstra,
  et~al.]{vinyals2016matching}
Oriol Vinyals, Charles Blundell, Tim Lillicrap, Daan Wierstra, et~al.
\newblock Matching networks for one shot learning.
\newblock In \emph{Advances in Neural Information Processing Systems}, pages
  3630--3638, 2016.

\bibitem[Wang et~al.(2016)Wang, Kurth-Nelson, Tirumala, Soyer, Leibo, Munos,
  Blundell, Kumaran, and Botvinick]{wang2016learning}
Jane~X Wang, Zeb Kurth-Nelson, Dhruva Tirumala, Hubert Soyer, Joel~Z Leibo,
  Remi Munos, Charles Blundell, Dharshan Kumaran, and Matt Botvinick.
\newblock Learning to reinforcement learn.
\newblock \emph{arXiv preprint arXiv:1611.05763}, 2016.

\bibitem[Weerts et~al.(2018)Weerts, Meuller, and Vanschoren]{Weerts2018}
H.~Weerts, M.~Meuller, and J.~Vanschoren.
\newblock Importance of tuning hyperparameters of machine learning algorithms.
\newblock Technical report, TU Eindhoven, 2018.

\bibitem[Wever et~al.(2018)Wever, Mohr, and H{\"u}llermeier]{weverml}
Marcel Wever, Felix Mohr, and Eyke H{\"u}llermeier.
\newblock Ml-plan for unlimited-length machine learning pipelines.
\newblock In \emph{AutoML Workshop at ICML 2018}, 2018.

\bibitem[Wistuba et~al.(2015{\natexlab{a}})Wistuba, Schilling, and
  Schmidt-Thieme]{Wistuba2015}
M.~Wistuba, N.~Schilling, and L.~Schmidt-Thieme.
\newblock Learning hyperparameter optimization initializations.
\newblock In \emph{2015 IEEE International Conference on Data Science and
  Advanced Analytics (DSAA)}, pages 1--10, 2015{\natexlab{a}}.

\bibitem[Wistuba et~al.(2015{\natexlab{b}})Wistuba, Schilling, and
  Schmidt-Thieme]{Wistuba2015b}
M.~Wistuba, N.~Schilling, and L.~Schmidt-Thieme.
\newblock Hyperparameter search space pruning, a new component for sequential
  model-based hyperparameter optimization.
\newblock In \emph{ECML PKDD 2015}, pages 104--119, 2015{\natexlab{b}}.

\bibitem[Wistuba et~al.(2018)Wistuba, Schilling, and
  Schmidt-Thieme]{Wistuba2018}
Martin Wistuba, Nicolas Schilling, and Lars Schmidt-Thieme.
\newblock Scalable {G}aussian process-based transfer surrogates for
  hyperparameter optimization.
\newblock \emph{Machine Learning}, 107\penalty0 (1):\penalty0 43--78, 2018.

\bibitem[Wolpert and Macready(1996)]{wolpert+96}
D.H. Wolpert and W.G. Macready.
\newblock No free lunch theorems for search.
\newblock Technical Report SFI-TR-95-02-010, The Santa Fe Institute, 1996.

\bibitem[Yang et~al.(2018)Yang, Akimoto, Kim, and Udell]{yang2018}
C.~Yang, Y.~Akimoto, D.W Kim, and M.~Udell.
\newblock Oboe: Collaborative filtering for automl initialization.
\newblock \emph{arXiv preprint arXiv:1808.03233}, 2018.

\bibitem[Yogatama and Mann(2014)]{yogatama2014efficient}
Dani Yogatama and Gideon Mann.
\newblock Efficient transfer learning method for automatic hyperparameter
  tuning.
\newblock In \emph{AI and Statistics}, pages 1077--1085, 2014.

\bibitem[Yosinski et~al.(2014)Yosinski, Clune, Bengio, and
  Lipson]{yosinski2014transferable}
Jason Yosinski, Jeff Clune, Yoshua Bengio, and Hod Lipson.
\newblock How transferable are features in deep neural networks?
\newblock In \emph{Advances in neural information processing systems}, pages
  3320--3328, 2014.

\end{thebibliography}

\end{document}